# Mapping-to-Parameter Nonlinear Functional Regression with Novel B-spline Free Knot Placement Algorithm

Chengdong Shi, Ching-Hsun Tseng, Wei Zhao, Xiao-Jun Zeng*

Department of Computer Science, University of Manchester

**x.zeng@manchester.ac.uk*

***Abstract***

**We propose a novel approach to nonlinear functional regression, called the Mapping-to-Parameter function model, which addresses complex and nonlinear functional regression problems in parameter space by employing any supervised learning technique. Central to this model is the mapping of function data from an infinite-dimensional function space to a finite-dimensional parameter space. This is accomplished by concurrently approximating multiple functions with a common set of B-spline basis functions by any chosen order, with their knot distribution determined by the Iterative Local Placement Algorithm, a newly proposed free knot placement algorithm. In contrast to the conventional equidistant knot placement strategy that uniformly distributes knot locations based on a predefined number of knots, our proposed algorithms determine knot location according to the local complexity of the input or output functions. The performance of our knot placement algorithms is shown to be robust in both single-function approximation and multiple-function approximation contexts. Furthermore, the effectiveness and advantage of the proposed prediction model in handling both function-on-scalar regression and function-on-function regression problems are demonstrated through several real data applications, in comparison with four groups of state-of-the-art methods.**

## 1. Introduction

Advancements in scientific measurement systems have led to increased collection of data that are observed repeatedly across a continuum, such as time or space, in various applications areas, including health monitoring (Shang and Hyndman 2017), financial markets (Das et al. 2019), and meteorology (Wang et. al 2020). Such data can be characterized by one or more functions, known as functional data. The primary attributes that distinguish other data categories are their intrinsically infinite-dimensional nature and generation by smooth underlying processes. While their infinite dimensionality configuration offers rich information for academic investigation, it renders standard data analysis techniques inapplicable without modification. A common way to handle functional data is to discretize the functions into vectors, and then process them through traditional multivariate data analysis methods. However, this discretization strategy may result in a high dimension of the input space as well as a failure in fully exploiting the inherent smooth functional behavior underlying the data. Consequently, the field of Functional Data Analysis (FDA) has garnered substantial research attention (e.g., Ramsay and Silverman 2005; Ferraty and Vieu 2006; Wang et al. 2016). FDA stands out by treating each function as an individual sample element and employing tailored techniques to analyze functional data for specific tasks, including Functional Regression (Morris, 2015), Classification (Preda et.al 2007), or Clustering (Jacques and Preda 2014).

Functional Regression encompasses three categories: 1) Scalar-on-Function (SOF) with function input and scalar output; 2) Function-on-Scalar (FOS) with scalar input and function output; 3) Function-on-Function (FOF) with both function input and output. While linear models for these regression problems have been extensively investigated, limited research work has been dedicated to the nonlinear model, particularly concerning the FOF regression problem.

Within the existing literature, the nonlinear FOF regression problem has been approached through two main avenues. The first approach is a statistics-based methodology that follows a two-step procedure. It begins by representing functional data through basis functions expansion and subsequently estimating the nonlinear relationship between functional predictor and functional response using statistical methods. The second approach, known as Functional Neuron Network (FNN), is a data-driven method that leverages the capabilities of neural networks to capture and learn the complex nonlinear mapping between functional inputs and outputs in an end-to-end manner.

In statistical methodologies, the initial step involves employing basis function expansion to represent functional data with the coordinates of their projection on a well-chosen base while preserving sufficient accuracy. Several families of basis functions have been investigated in the FDA literature, including B-splines (Beyaztas and Shang 2022, Luo and Qi 2017), Fourier (Thinda et a. 2023; Luo and Qi 2019), and Wavelet (Luo and Qi 2021). Functional Principal Components (FPCs) serve as analogs to basis functions and are derived empirically from the eigenfunctions of the covariance operator of functional data (Kowal and Bourgeois, 2020; Yao and Müller 2005). Among the available options, B-spline basis functions are often preferred due to their superior flexibility in capturing highly nonlinear complex patterns. A B-spline has a piecewise polynomial structure where the polynomial pieces are connected at knots. The accuracy of a B-spline approximation of a function is significantly impacted by both the location and the number of knots. Despite their potential, the application of B-spline in current FDA research remains oversimplified and less effective, with most studies relying on a uniform knot distribution with a predetermined knot number, suitable only for homogeneous curves (Jupp 1978).

Substantial efforts have been dedicated to the investigation of the optimal knot placement in non-uniform spaces within the field of function approximation. Various approaches for free knot placement have emerged, including iterative knot placement such as knot insertion (Tjahjowidodo et al. 2017; Tjahjowidodo et al. 2015) and knot removal method (Kang et al. 2015; Conti et al. 2001), stochastic methods like Genetic Algorithms (Goldenthal and Bercovier 2004) and Particle Swarm Optimization ( Gálvez and Iglesias 2011), and heuristic methods based on specific properties of the input dataset such as derivative (Yeh et al. 2020), curvature (Liang et al. 2017) and arc-length (Michel and Zidna 2021). These existing free knot placement techniques have consistently exhibited superior performance in approximating a single function compared to the equidistant knot placement method. However, their direct implementation in the FDA context is limited by the fact that the primary concern of functional representation in the FDA lies in approximating a set of functions rather than a single function.

The subsequent step after functional representation involves modeling the functional relationship between functional predictors and responses. In the case of the standard functional linear regression (FLR), a linear parametric form is assumed for this relationship:

$$Y(s) = \beta_o(s) + \int X(t)\beta(s,t)dt + \varepsilon(s) \tag{1}$$

where $X(t)$ and $Y(s)$ denote the functional predictor and response, $\beta(s,t)$ represents the bivariate coefficient function, and $\varepsilon(s)$signifies the random error function (Luo and Qi 2017; Luo and Qi 2021; Beyaztas and Shang 2022). To alleviate the linearity assumption in functional modeling, Ferraty and Vieu (2006) investigated fully nonparametric function models based on kernel smoothing techniques. However, selecting an appropriate metric for the kernel in the function space remains challenging, as highlighted by Delaige and Hall (2010), due to the inherent complexity of the function space itself. Alternatively, several methods have introduced structural constraints to the model, allowing for a degree of flexibility while restricting the learning to only certain types of nonlinear relations. For example, the functional quadratic model, as studied by Sun and Wang (2020), assumes a quadratic relationship between the functional predictor and response by incorporating a quadratic term. Another well-known instance is the functional additive model (FAM), explored by Maller, Wu, and Yao (2013), which assumes an additive effect of time. In general, the current statistical approach in the FDA has been dominated by functional modeling within infinite-dimensional spaces. This attribute poses challenges in developing nonlinear functional regression models, given that the explicit expression for the nonlinear functional mapping between predictor and response in infinite-dimensional spaces remains elusive owing to the intricate nature of the functional spaces.

Unlike statistical approaches that manually identify nonlinear functional relationships between functional predictor and response, the alternative approach, Functional Neuron

Network (FNN), can automatically interrogate these nonlinear functional relationships through neural networks based on the data itself. Pioneering works by Rossi and Conan-Guez (2005) and Ross et al. (2005) first introduced the concept of FNN, with an extension of the traditional neuron network architecture by transforming the first hidden layer into a functional layer with functional neurons capable of handling functional inputs. The exploration of FNN models has led to various modifications in the field. Thind et al. (2022) proposed the FuncNN model, which goes beyond the traditional FNN architecture by enabling the incorporation of both functional and scalar inputs. Furthermore, Yao, Mueller, and Wang (2021) developed the AdaFNN model, which introduces a basis layer with each neuron acting as a micro neural network, allowing for adaptive learning of optimal basis functions specific to the functional data. While these advancements have primarily focused on SOF regression, a limited number of studies have addressed the FOF regression problem by transforming the output layer into a functional layer with functional neurons to deliver a final functional estimation (Rao and Reimher 2023, and Hsieh et al. 2021). In contrast to statistical methods consisting of two separate steps, the FNN operates in an end-to-end fashion. In fact, however, the first hidden layer in FNNs serves the same purpose as the function representation in statistical methods, with the number of neurons determining the number of basis functions used to approximate functional input. Nevertheless, the FNN model relies on a trial-and-error approach to determine the optimal number of neurons, which could be time-consuming and lack a theoretical guide. In addition, the learning process of FNNs proceeds still in the function space, where the nonlinear relationship is learned through iterative adjustment of the functional weights. This leads to the fact that two drawbacks commonly associated with neural networks, high computational cost and lack of interpretability, can be further amplified in FNNs due to the complex nature of the function space.

In this paper, we present a novel methodology called the Mapping-to-Parameter (M2P) nonlinear FOF regression model. This model comprises two procedures: (1) representing the input and output functions using B-spline basis functions expansion and mapping them into finite-dimensional parameter space, and (2) learning the nonlinear functional relationship between these input and output functions in the parameter space using usual supervised learning techniques. The significant advancement of our model is its capability to overcome obstacles encountered by existing approaches in uncovering unknown nonlinear functional relationships in complex function spaces, by shifting the learning process into the parameter space. Furthermore, we introduce a pathway for approximating a set of functions by determining a common set of B-spline basis functions capable of representing all given functions with the required precision. For this purpose, we propose a new free knot placement algorithm, Iterative Local Placement (ILP). The ILP algorithm aims to identify the optimal common knot distribution that closely resembles the trajectory of the given functions considering the maximum local complexity among all functions. This ensures the approximation embodies the key characteristics of each function while preserving the intended level of accuracy. The main contribution of this work includes:

(i) ***Functional Representation with the Novel Knot Placement Algorithm*:** The proposed ILP algorithm delivers a theoretical guide for determining the optimal common set of B-spline basis functions for input or output functions based on the characteristics of the data itself. Functional representation based on the ILP algorithm marks the pioneering integration of free knot placement techniques within the scope of functional regression.

(ii) ***Uncovering Nonlinear Relationship in Parameter Space via M2P model***: The introduction of our novel M2P model facilitates the learning of the nonlinear mapping between input and output functions within parameter space. This framework is not restricted to neural network architecture alone but can also be seamlessly integrated with any conventional supervised learning techniques. The reduced dimensionality within the parameter space, as compared to the function or observed discrete real-valued space, contributes to a substantial cutback in computational expenses required.

(iii) ***Real-world Evaluation of Approximation Performance of the ILP algorithm***: An in-depth comparative study has been conducted to evaluate the approximation performance of the proposed ILP algorithms against the equidistant knot placement method under both single and multiple-function approximating settings. This empirical analysis demonstrates the efficacy of our proposed algorithms.

(iv) ***Validation and Benchmarking of Predictive Performance of the M2P Model:*** The robustness and effectiveness of our M2P model are assessed through its employment in handling SOF and FOF regression tasks, leveraging a variety of real-world examples. By benchmarking against several widely used prediction models, we provide empirical evidence affirming the superior performance of the M2P model.

The organization of this paper is as follows: Section 2 outlines the two primary problems of this study. Section 3 presents the proposed free knot placement algorithm, followed by a description of the M2P function model in Section 4. Section 5 offers experimental results for the proposed free knot algorithm in single and multiple-function approximation contents and presents the prediction performance of the M2P model in both settings of Scalar Output Prediction and Functional Output Prediction. The final section provides a summary and proposes future research directions in the field.

## 2. Problem Formulation

In this section, we review the basics of B-spline theory, introduce the essential notations used in this work, and subsequently formalize the two core problems addressed in this study.

Let $f \in C([T])$ be a function defined over an interval $T = [a, b]$, and for integers $n$ and $p$ where $n > p \geq 0$, consider a knot sequence $\boldsymbol{\xi}$:

$$\boldsymbol{\xi} \coloneqq \{\xi_j\}_{j=1}^{n+p+1} = \{\xi_1 \leq \xi_2 \leq \cdots \leq \xi_{n+p+1}\}, n \in \mathbb{N}, p \in \mathbb{N}_0 \tag{2}$$

A degree $p$ B-spline approximation of the function $f$ can be written as:

$$\hat{f}(t) \coloneqq \sum_{q=1}^{n} B_{q,p,\xi}(t)\, \beta_q, \ \ t \in [\xi_1, \xi_{n+p+1}] \tag{3}$$

where $\hat{f}(t)$ is the fitted function at point $t$, $\beta_q$ is the coefficient for the $q$-th basis function for $\hat{f}$, and $B_{q,n,\xi}$ is the $q$th B-spline basis functions of degree $p$ defined over the knot sequence $\boldsymbol{\xi}$. It is assumed that the knot sequence $\boldsymbol{\xi}$ is open with multiplicity $p+1$ at the interval $[a, b]$, i.e.

$$a \coloneqq \xi_1 = \cdots = \xi_{p+1} < \xi_{p+2} \leq \cdots \leq \xi_n < \xi_{n+1} = \cdots = \xi_{n+p+1} \eqqcolon b \tag{4}$$

The B-spline basis function $B_{q,p,\xi}(t)$ is recursively defined as:

$$\begin{gathered} B_{q,0,\xi}(t) \coloneqq \begin{cases} 1, & if\ t \in [\xi_q, \xi_{q+1}) \\ 0, & otherwsie \end{cases} \\ B_{q,p,\xi}(t) \coloneqq \frac{t-\xi_q}{\xi_{q+p}-\xi_q} B_{q,p-1,\xi}(t) + \frac{\xi_{q+p+1}-t}{\xi_{q+p+1}-\xi_{q+1}} B_{q+1,p-1,\xi}(t) \end{gathered} \tag{5}$$

Based on the recurrence relation (4), a B-spline exhibits the following key properties:

a. *Local support.* B-spline has local support over $p+1$ knot spans, i.e.,

$$B_{q,p,\xi}(t) = 0, t \notin [\xi_q, \xi_{q+p+1}) \tag{6}$$

*b.* *Nonnegativity.* B-spline is nonnegative everywhere, and positive within its support.

*c.* *Local partition of unity:* According to the local Marsden identify (Lyche et.al 2018), for $p+1 \leq m \leq n$,

$$\sum_{q=m-p}^{m} B_{q,p,\xi}(t) = 1, t \in [\xi_m, \xi_{m+1}) \tag{7}$$

Similar to classical function approximation theories, B-spline approximation primarily deals with the single-function approximation problem. We now introduce a novel type of approximation problem, named the *Unified Multifunction Approximation*:

*Definition*: *Unified Multifunction Approximation*

*Given a set of N functions* $F = \{f_i : f_i \in C(T), i = 1, \dots, N\}$, *the objective to identify a common set of* $n$ *basis functions* $\{\phi_1, \phi_2, \dots, \phi_n\}$, *such that for each* $f_i$ *in* $F$, *there exists a set of coefficients* $\{\beta_{i,q} \in \mathbb{R}\}_{q=1}^{n}$ *for which the approximation:*

$$\hat{f}_i(t) = \sum\nolimits_{q=1}^{n} \phi_q(t)\, \beta_{i,q} \tag{8}$$

*satisfies* $\|\hat{f}_i(t) - f_i(t)\| < \varepsilon$ *for all* $t \in T$, *and for all* $i = 1, \dots, N$, *where* $\varepsilon$ *is some error* *tolerance and* $\|\cdot\|$ *represents a norm defined in a Hilbert space.*

In our study, we consider *N* subjects, each with a continuous input function $X_i(t) \in L^2(T)$ over a compact interval *T*, and a corresponding continuous output function $Y_i(s) \in L^2(S)$ over a compact interval *S*, where $L^2$ denotes a Hilbert space. In practical scenarios, the continuous processes of both $X(t)$ and $Y(s)$ can be only measured discretely at points $\{t_j\}$ for $j = 1 \ldots, J$ and $\{s_r\}$ for $r = 1 \ldots, R$.

Our first problem involves approximating each input function $X_i(t)$ and output function $Y_i(s)$ respectively using common sets of B-spline basis functions that satisfy a predetermined error criterion. Determining separate common sets of B-spline basis functions for both input and output functions ensures that the essential information from each function of the corresponding variable is captured in the same coordinate system, thus facilitating effective learning in subsequent regression models.

Once the knot sequence for the common set of B-spline basis functions is determined, the coefficients corresponding to each function can be obtained by minimizing the least square error between the approximated function and the true function, e.g., for the approximation of $X_i(t)$:

$$\underset{\boldsymbol{\beta}}{argmin} \sum_{j=1}^{J} \left(\hat{X}_i(t_j) - X_i(t_j)\right)^2 \tag{9}$$

where $X_i(t_j)$ is the actual observed value of $X_i(t)$ at time point $t_j$

Hence, the first problem can be divided into two knot optimization challenges: to determine the optimal common knot placement $\boldsymbol{\xi}_X^*$ and $\boldsymbol{\xi}_Y^*$ for the input variable $X$ and the output variable $Y$, respectively, with the minimal number of knots such that both B-spline approximation $\hat{X}_i$ and $\hat{Y}_i$ satisfy the accuracy requirements

$$\|\hat{X}_i(t) - X_i(t)\| < \varepsilon^{(X)}, \qquad \|\hat{Y}_i(s) - Y_i(s)\| < \varepsilon^{(Y)} \tag{10}$$

for all $i$ and for all $t$ and $s$ within the domains of interest.

Our second problem is to estimate the unknown nonlinear functional mapping, $g : L^2(T) \to L^2(S)$ between the input function $X_i(t)$and the output function $Y_i(s)$ through supervised learning methods. The optimal functional approximator $g$ is determined by minimizing the mean squared error (MSE) loss function between the actual output $Y_i(s)$ and the predicted output $\tilde{Y}_i(s)$ across all subjects in an L2 sense:

$$\underset{g \in L^2}{argmin} \frac{1}{N} \sum_{i=1}^{N} \int_{s \in S} (Y_i(s) - \tilde{Y}_i(s))^2 \, ds \tag{11}$$

where $N$ is the total number of subjects and $\tilde{Y}_i(s) = g(X_i(t))$ is the predicted output function.

## 3. Iterative Local Knot Placement for B-spline approximation

This section introduces a novel free knot placement algorithm called the Iterative Local Placement (ILP) algorithm, which is used to determine a common knot distribution to form a common set of B-spline basis functions which can approximate a set of functions with the same desired level of approximation accuracy.

We first consider the problem of approximating a single function: let $f \in C(T)$ be a function on an interval $T = [a, b]$, the goal is to find a set of degree $p$ B-spline basis function $\{B_{q,p,\xi}\}_{q=1}^{n}$ defined on a knot sequence $\boldsymbol{\xi} = \{\xi_j\}_{j=1}^{n+p+1} \subseteq T$, such that the B-spline approximant $Qf$ of $f$, given by:

$$Qf(t) := \sum_{q=1}^{n} B_{q,p,\xi}(t) \, \beta_q, \;\; t \in [\xi_1, \xi_{n+p+1}] \tag{12}$$

with suitable coefficients $\beta_q$, satisfies

$$|f(t) - Qf(t)| < \varepsilon \tag{13}$$

for all $t \in T$ and a given tolerance $\varepsilon$.

The local support property in (6) of B-splines implies that the value of $Qf$ at a point $t$ depends only on the values of $f$ in a local neighborhood of that point. Specifically, for $p + 1 \leq m \leq n$,

$$\sum_{q=1}^{n} B_{q,p,\xi}(t)\,\beta_q = \sum_{q=m-p}^{m} B_{q,p,\xi}(t)\,\beta_q,\ \ t \in [\xi_m, \xi_{m+1}] \tag{14}$$

This property motivates a localized approach to construct a B-spline approximation with the desired accuracy by controlling the local approximation error over each knot span.

The Lagrange Error Theorem (Shadrin 1995) gives an upper bound for Tayler polynomial approximation error. An extension of this theorem to B-spline spaces is provided in Theorem 32 of Lyche et.al (2018), providing a local error bound for the B-spline approximation over a knot span. Based on these theorems, we hereby present a lemma on a local and a global upper bound of B-spline approximation error in a $L_\infty$ norm.

*Lemma 1: Let* $f \in C^{p+1}(T)$ *be a function that is* $p+1$ *times continuously differentiable on a close interval T, and let* $Qf$ *be a degree* p *B-spline approximation to* $f$ *defined over the knot sequence* $\boldsymbol{\xi} = \{\xi_j\}_{j=1}^{n+p+1} \subseteq T$. *For any* $m \in [p+1, n]$, *the local error bound for* $Qf$ *in the* $L_\infty$*-norm over the knot span* $[\xi_m, \xi_{m+1}]$ *is given by*

$$\|f - Qf\|_{L_\infty([\xi_m,\xi_{m+1}])} = \sup_{t\in[\xi_m,\xi_{m+1}]} |f(t) - Qf(t)| \le \frac{h_m^{p+1}}{p!} \|D^{p+1}f\|_{L_\infty([\xi_m,\xi_{m+1}])} \tag{15}$$

*where* $h_m = \xi_{m+1} - \xi_m$ *is the length of the m-th knot interval and* $\|D^{p+1}f\|_{L_\infty([\xi_m,\xi_{m+1}])} := \max_{\xi_m \le t \le \xi_{m+1}} |f^{(p+1)}(t)|$ *with* $f^{(p+1)}(t)$ *denoting the* $(p+1)$*-th derivative of* $f$ *evaluated at t.*

*If there is a constant* $\varepsilon > 0$ *such that for all* $m \in [p+1, n]$*:*

$$\frac{h_m^{p+1}}{p!} \|D^{p+1}f\|_{L_\infty([\xi_m,\xi_{m+1}])} < \varepsilon \tag{16}$$

*then the* $L_\infty$ *global error of the B-spline approximation* $Qf$ *to* $f$ *is bounded by* $\varepsilon$*, i.e.,*

$$\|f - Qf\|_{L_\infty(T)} < \varepsilon \tag{17}$$

Proof of Lemma 1 is provided in Section 1.1 of the supplementary document.

We now extend the single function approximation problem to the unified multifunctions approximation problem: given a set of $N$ functions $F = \{f_i(t) : f_i \in C(T), i = 1, \dots, N\}$ over the same interval $T = [a, b]$, the goal is to find a common set of degree $p$ B-spline basis function $\{B^*_{q,p,\xi}\}_{q=1}^{n}$ defined on a knot sequence $\boldsymbol{\xi}^* = \{\xi_j^*\}_{j=1}^{n+p+1} \subseteq T$, whose linear expansion approximant $\{Qf_i\}_i^N$ for each respective function $f$ such that

$$|f_i(t) - Qf_i(t)| < \varepsilon \tag{18}$$

for all $t \in T$ and all $i$.

To solve this new function approximation problem, we adopt a strategy considering the "worst-case scenario". Specifically, if a set of B-spline basis functions can accommodate the function with the highest degree of complexity among all given functions, it can also handle each simpler individual case within the same error bound. Here, the complexity of a function is quantified by the magnitude of its (p + 1)-derivative, since the derivative signifies the rate of change of a function. A higher derivative value at a specific point $t \in T$ corresponds to higher variability and complexity of the function at that location. Consequently, we characterize the most complex dynamics of a given set of functions by considering the maximum value of the ( p+ 1)-derivative across all given functions at each point $t_j \in T$. Under this setting, We apply *Lemmas 1* to the case of the unified multifunction approximations problem, and we have the following theorem:

*Theorem 1:*

*Let $\{f_i \in C^{n+1}(T)\}_{i=1}^{N}$ be a set of N functions defined over the same interval $T$. And let $\{B^*_{q,p,\xi}\}_{q=1}^{n}$ be a set of degree p B-spline basis functions defined upon a knot sequence $\boldsymbol{\xi}^* = \{\xi_j^*\}_{j=1}^{n+p+1} \subseteq T$. Supposed that for each t $\in T$, the (p+1)-th derivatives of the function $f_i$ are uniformly bounded, i.e., there exists a constant $C_t$, such that*

$$\max_{i \in [1,N]} \left| f_i^{(p+1)}(t) \right| \le C_t, \qquad \forall t \in T \tag{19}$$

*If there exists a constant $\varepsilon > 0$ such that for all $m \in [p + 1, n]$,*

$$\frac{\max\limits_{\xi_m^* \le t \le \xi_{m+1}^*} |C_t|}{p!} {h_m^*}^{p+1} < \varepsilon \tag{20}$$

*where* ${h_m^*}^{p+1} = \xi_{m+1}^* - \xi_m^*$ *is the length of the m-th knot interval,*

*then each function* $f_i$ *is approximated by the B-spline basis function set* $\{B_{q,p,\xi}^*\}_{q=1}^n$ *such that the global approximation error in the* $L_\infty$ *norm for each i satisfies:*

$$\|f_i - Qf_i\|_{L_\infty(T)} < \varepsilon \tag{21}$$

*where* $Qf_i$ *denotes the B-spline approximant to* $f_i$

Proof of Theorem 1 is provided in Section 1.2 of the supplementary document.

Theorem 1 asserts that if one chooses a set of B-spline basis functions with an appropriate knot sequence such that the local approximation errors for all $f_i$ on each span of knots are bounded by ε in the $L_\infty$ norm, then it follows that the global $L_\infty$ errors for all these approximations are also bounded by $\varepsilon$. Such insight serves as a guiding principle when applying B-splines for unified multifunction approximation in practice. By trying to regulate the local approximation error during the knot selection process, the theorem guarantees that the desired global approximation error is achieved for all functions.

The computation of derivatives from discretely observed data of an unknown function demands careful consideration due to the potential amplification of noise. Although this is not our principal area of interest, several solutions have been put forth to mitigate this challenge (Breugel, Kutz and Brunton, 2020). For our present objectives, we opt to employ the Forward Finite Difference Approximation method. Specifically, the derivative of a function $f$ at the point $t$ is computed as follows:

$$f'(t) = \lim_{\Delta t \to 0} \frac{f(t + \Delta t) - f(t)}{\Delta t} \tag{22}$$

Higher-order derivatives can be computed analogously.

Based on the preceding theoretical analysis and groundwork, we now present the core procedures of the ILP algorithm for unified multifunction approximations. This concise representation is summarized in Algorithm 1.

1. Calculate the *(p+1)*-th derivatives for all function $f_i(t)$ at each $t$ in $T$

2. Determine the maximum value of these *(p+1)*-th derivatives across all given functions at each $t$.

3. Begin with no knot in the knot sequence.

4. Find the longest knot span, from left to right, where the local approximation error over such knot span meets the bound specified in Equation (20).

5. Position the knots at the boundaries of the identified subinterval.

6. Repeat the iterative process until the last subinterval satisfies the specific criterion.

---

**Algorithm: The Iterative Local Placement Algorithm**

---

***Input:*** *A set of N sequences of discrete observation* $\boldsymbol{X}_i \in \mathbb{R}^J$, *corresponds to the unknown functions* $f_i$ *at discrete timesteps* $t_j$ *where* $a = t_1 < t_2 < \cdots < t_J = b$ *and* $i = 1, \ldots, N$, *the degree of B-spline,* $p$ *and error tolerance,* $\varepsilon$

***Output:*** *Interior knot sequence* $\boldsymbol{\xi}^* = \{\xi_1^*, \ldots\}$

***Initialization of variables:*** *assign 1 to variable m, a to variable* $\xi_0^*$

*1*    *For each j=1,…,J and for each i=1,…,N, compute* $f_i^{(p+1)}(t_j)$

     *For each* $t_j$, *compute* $C_{t_j} = \max_{i\in[1,N]} \left| f_i^{(n+1)}(t_j) \right|$

*2*    ***While*** $m < J$ ***do the following***

*3*      ***for*** $t_q$ ***in*** $[\xi_{m-1}^*, b]$*:*

       ***a.*** *Compute* $M_{max} \leftarrow \max_{\xi_{m-1}^* \le t_j \le t_q} \left| C_{t_j} \right|$;

       ***b.*** *set* $\delta \leftarrow \frac{M_{max}}{p!} \left| t_q - \xi_{m-1}^* \right|^{p+1}$

       ***c.*** ***If*** $\delta > \varepsilon$ ***, then break.***

     ***end for***

*4*      $t_q \leftarrow k_\xi^*$ *&* $m \leftarrow m + 1$

*5*      ***if*** $t_q = b$, ***then break.***

   ***end while***

***return*** $\boldsymbol{\xi}^* = \{\xi_1^*, \ldots\}$

---

Figure 1 illustrates an example where a real-world solar irradiance data sequence, collected over a single day at minute intervals, is being approximated by a set of linear B-spline basis functions, with the knot locations determined via the equidistant knot placement method (top) and the ILP algorithm (bottom). The same number of knots is used in both cases. For the equidistant knot strategy, underfitting is observed in areas of fluctuation since there are not enough knots allocated in the area to capture the complexity, while too many knots are located in the smooth region and cause overfitting. The ILP_pw algorithm addresses these issues by increasing the density of knots in higher fluctuating areas, and conversely decreasing knot density in smoother regions.

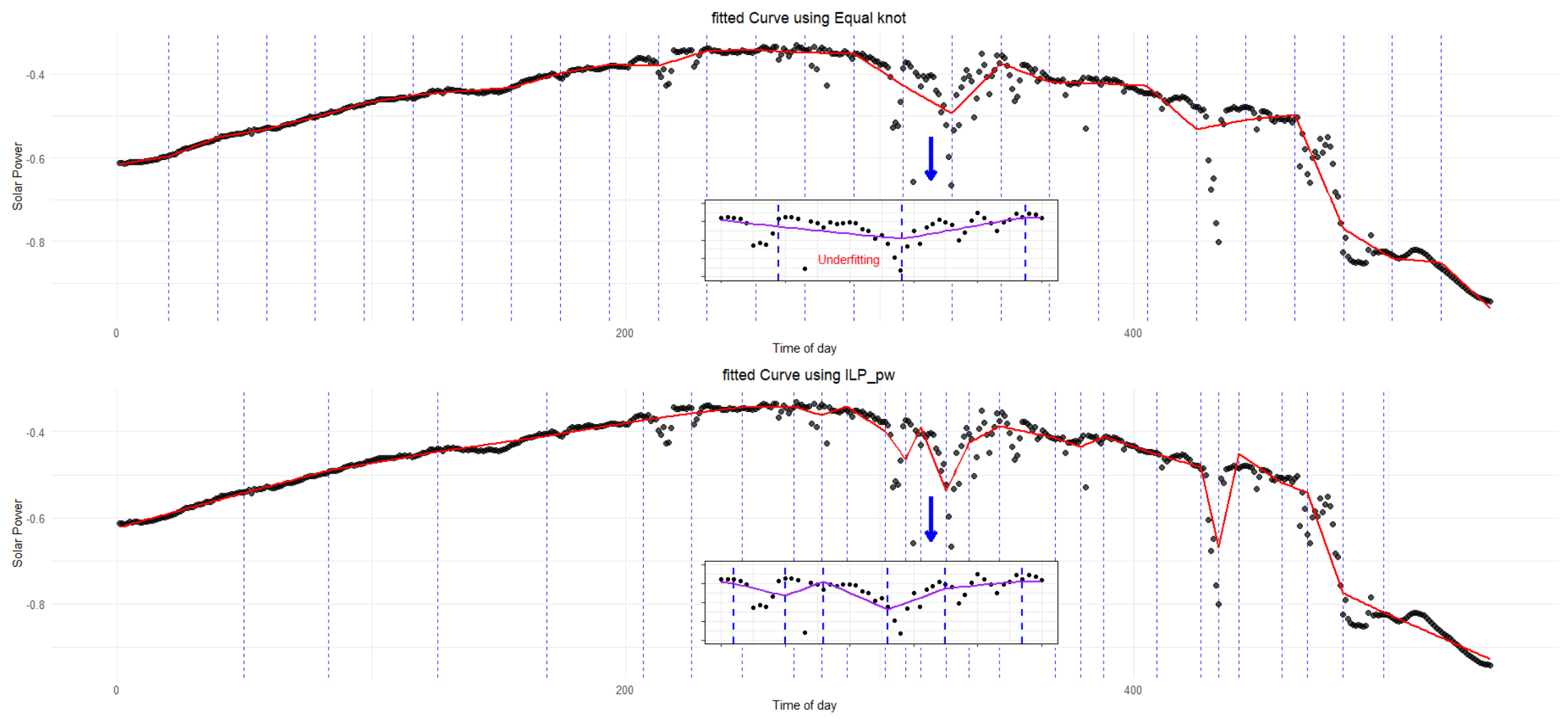

Figure 1. Demonstration of ILP and the equidistant knot placement algorithm. The approximated curve is red. The knots used are indicated by the blue vertical dashed lines.

## 4. Mapping-to-Parameter Function-on-Function Regression Model

In this section, we introduce our proposed model framework for a nonlinear FOF regression model, namely the Mapping-to-Parameter (M2P) functional regression model. The main idea of the M2P functional model is to map functional data from an infinite-dimensional function space into a finite-dimensional parameter space and learn the regression model in the parameter space using any usual supervised learning techniques. This framework was initially introduced in a previous work (Shi and Zeng, 2022) but was limited to the SOF model. Figure 3 presents the overall architecture of the M2P model.

The main procedures of the M2P functional model are as follows:

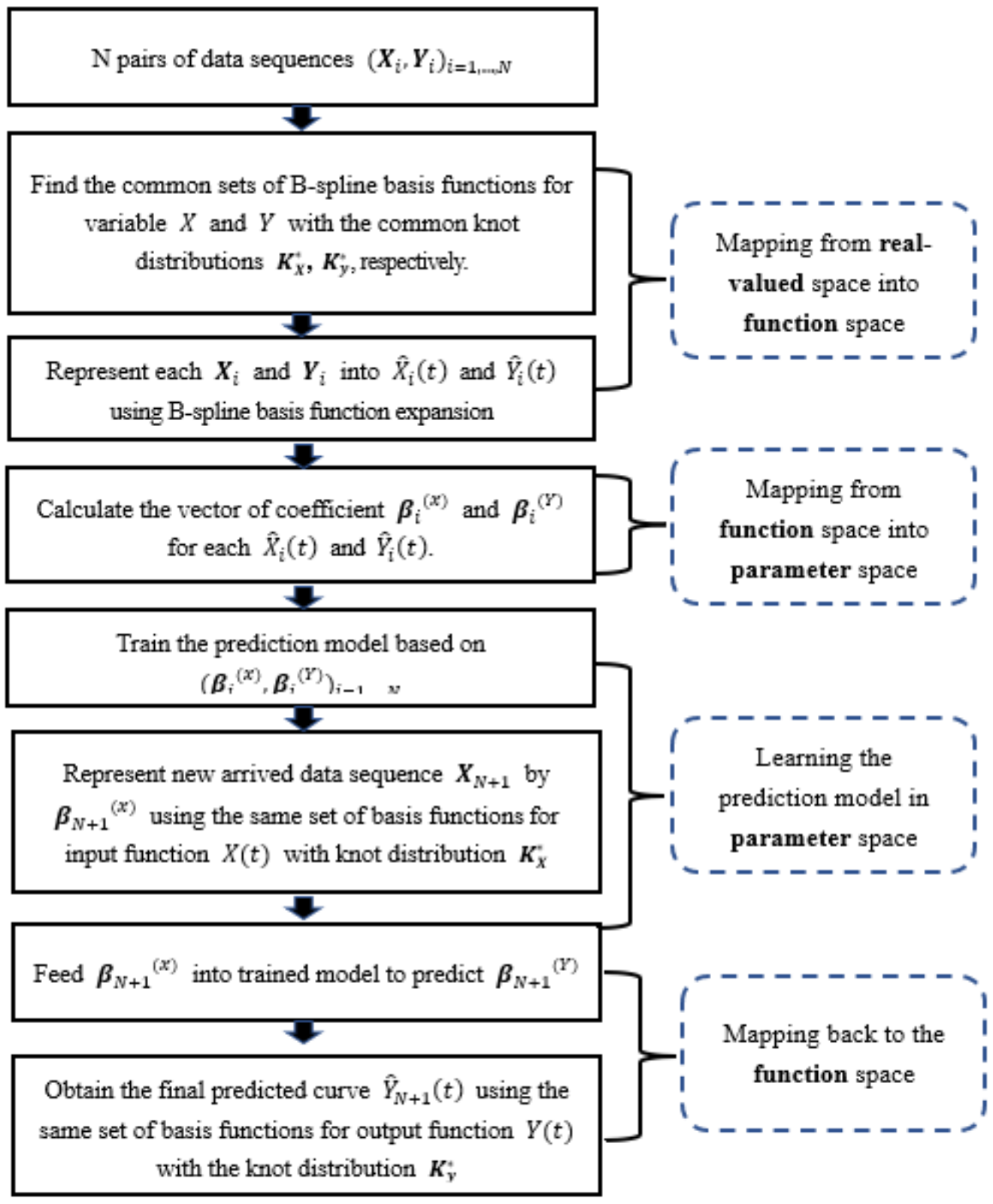

Figure 2. The schematic diagram of the proposed M2P model

### *i. Mapping from observed discrete real-valued space into function space (i.e., functional data representation)*

In practice, for N pairs of functional data $\left(X_i(t), Y_i(s)\right)_{i=1,\ldots,N}$, their observations are usually only available at discrete time points $\{t_j\}_{j=1}^{J}$ and $\{s_r\}_{r=1}^{R}$, represented as $(\boldsymbol{X}_i \in \mathbb{R}^J, \boldsymbol{Y}_i \in \mathbb{R}^R)_{i=1,\ldots,N}$. The initial procedure is to approximate each $\boldsymbol{X}_i$ and $\boldsymbol{Y}_i$ into B-spline approximation, $\hat{X}_i(t)$ and $\hat{Y}_i(t)$ in the form of Equation (3) based on the common knot distributions $\boldsymbol{\xi}_X^*$ and $\boldsymbol{\xi}_y^*$ for input and output, respectively. The ILP algorithm is employed to obtain the functional data representation with the optimal knot distribution.

### *ii. Mapping from functional space into parameter space*

To address the infinite-dimensional challenge, the fitted functions $\hat{X}_i(t)$ and $\hat{Y}_i(t)$ are represented in parameter space using the vector of their coefficients $\boldsymbol{\beta}_i^{(X)} \in \mathbb{R}^{n_x}$ and $\boldsymbol{\beta}_i^{(Y)} \in$ $\mathbb{R}^{n_y}$, respectively. The dimension of these coefficients corresponds to the number of basis functions used for the approximation, which is equal to the number of interior knots used in addition to the order of the fitted B-spline. Hence, for any input function $X_i$, the dimension of the coefficient vectors is identical for all $i$, owing to the use of the same set of B-spline basis functions for the approximation. This is also true for any output function $Y_i$. The coefficient values are calculated by minimizing the $L_2$ norm of the fitting error, as presented in Equation (9).

***iii. Learning the prediction model in the parameter space and mapping the result back to the function space***

A functional regression problem is subsequently reduced into a standard multiple multivariate regression problem with an input dimension $n_x$ and an output dimension $n_y$. The prediction model is trained based on $(\boldsymbol{\beta}_i^{(X)}, \boldsymbol{\beta}_i^{(Y)})_{i=1,\ldots,N}$ using any conventional supervisor learning algorithm. For the newly arrived data $\boldsymbol{X}_{N+1} = \{X_{N+1,1}, \ldots, X_{N+1,J}\}$ it needs to be represented to its functional form using the identical set of basis functions for input functions with the same knot distribution $\boldsymbol{\xi}_X^*$, and then obtains its corresponding coefficient into $\boldsymbol{\beta}_{N+1}^{(X)}$. This allows the data to be input into the trained model for the prediction of $\boldsymbol{\beta}_{N+1}^{(Y)}$. Finally, the targeted curve $\hat{Y}_{N+1}(t)$ can be determined in the form of Equation (3) using the same set of basis functions for output functions with the same knot distribution $\boldsymbol{\xi}_y^*$.

## 5. Experiment Results:

This section contains two main subsections: the first subsection evaluates the approximation performance for our proposed ILP algorithm in both contexts of single function approximation and unified multifunction approximation, compared to the standard equidistant knot placement strategy; the second subsection demonstrates the predictive performance of our proposed M2P functional regression model compared with a wide range of state-of-the-art techniques in handling several different real-world nonlinear SOF and FOF regression problems.

### *5.1. Function Approximating*

In this subsection, we concentrate on a solar irradiance dataset for the year 2016, sourced from the National Renewable Energy Laboratory Solar Radiation Database (NSRDB) (Andreas and Stoffel, 1981). The dataset is selected for the true solar time span from 8:00 and 16:00 and is normalized by a linear scaling method to lie within the boundaries of $[-1,1]$. Our investigation includes two approximation tasks using this dataset. The first task, single function approximation, seeks to estimate the solar irradiance curve for a particular day from its one-minute discrete observations. The second task, the unified multifunction approximation task, aims to identify a common set of B-spline basis functions that can effectively approximate the solar irradiance curve for all 366 days in 2016.

For each task, we evaluate the performance of our proposed ILP algorithm by comparing it to the equally spaced knot method. The approximation error between the true and approximated function is quantified using both Maximum Absolute Error (MaxAE) and Root Mean Squared Error (RMSE). The MaxAE computes the largest absolute discrepancy between the true and approximated functions:

$$MaxAE = \max_{j=1,\dots,J}\left(\left|f(t_j) - \hat{f}(t_j)\right|\right) \tag{23}$$

The RMSE, in contrast, calculates the square root of the mean of the squared difference between the true and approximated function values at all discrete time points across the entire domain:

$$RMSE = \sqrt{\frac{1}{J}\sum_{j=1}^{J}\left(f(t_j) - \hat{f}(t_j)\right)^2} \tag{24}$$

In the above, $f(t_j)$ and $\hat{f}(t_j)$ represent the actual and predicted values at $t_j$, $|\cdot|$ denotes the absolute value and $J$ is the total number of observed function values within the domain.

### *5.1.1. Single Function Approximation*

We conduct the single function approximation task separately for one representative day from each season of 2016, specifically January 1 (winter), March 20 (spring), June 21 (summer), and September 22 (fall). These representative days are selected using a standard K-means algorithm with four clusters, and their respective variation patterns are distinctively colored in Figure 4.

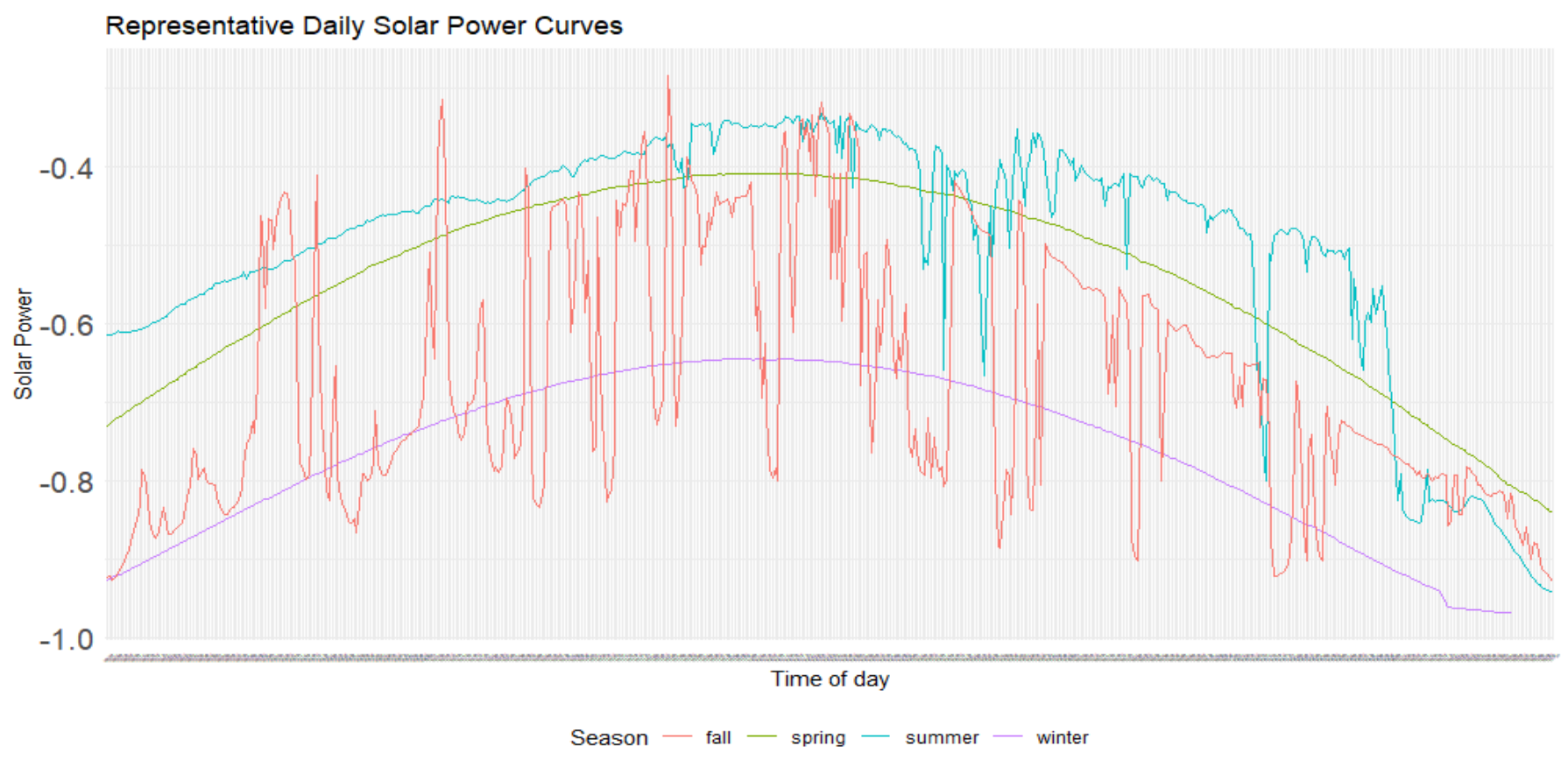

Figure 3. Daily Solar Pattern for four representative days in 2016

For a robust performance evaluation of our proposed ILP algorithm against the equally spaced knot method, we employ two comparison metrics. First, we contrast the approximation accuracies in terms of RMSE and MaxAE, while using an equal number of knots. Secondly, we compare the number of knots necessary to attain the same level of error. For the latter, we identify the optimal number of knots needed for each knot placement method, maintaining an identical level of Generalized Cross-Validation (GCV) Error level, a well-recognized criterion in function approximation that balances the trade-off between goodness-of-fit and model complexity.

The GCV error commonly provides an unbiased estimate of the prediction error by iteratively omitting one observation from the dataset, fitting the model to the remaining data, and then computing the prediction error for the omitted observation. The GCV Error is given by:

$$GCV = \frac{1}{J}\sum_{j=1}^{J}\left(f(t_j) - \hat{f}_{-j}(t_j)\right)^2 \tag{25}$$

where $f(t_j)$ denote the actual function values at $t_j$, and $\hat{f}_{-j}$ indicates the fit using all training sampling except for the $j$-th pair $(t_j,\ f_j)$.

Often, the GCV expression is simplified based on the smoothing matrix $\boldsymbol{S}$, represented as:

$$GCV = \frac{1}{J}\sum_{j=1}^{J}\left[\frac{f(t_j) - \hat{f}(t_j)}{1 - \boldsymbol{S}_{ii}}\right]^2 \tag{26}$$

In B-spline modeling, the smoothing matrix $S \in \mathbb{R}^{J\times J}$ operates as a projection mapping the observed data to the space spanned by the B-spline basis functions $\boldsymbol{\phi}$. It is defined as $\boldsymbol{S} = \boldsymbol{\phi}(\boldsymbol{\phi}^T\boldsymbol{\phi})^{-1}\boldsymbol{\phi}^T$ with $\boldsymbol{S}_{ii}$ referring to the i-th diagonal element of matrix $\boldsymbol{S}$. Here, $\boldsymbol{\phi} \in \mathbb{R}^{\boldsymbol{m}\times \boldsymbol{J}}$ signifies a matrix with each column corresponding to a B-spline basis function evaluated at all observation points, with $m$ as the total number of basis functions for $\hat{f}(t)$.

The selection of the order of the B-spline basis functions plays a pivotal role in B-spline approximation, significantly influencing the accuracy of the resulting approximation. This choice should be guided by the inherent complexity of the target function being approximated. While higher-order B-splines can capture more intricate and wiggly patterns in data, they risk amplifying oscillations and incurring overfitting, especially when approximating smoother functions. Conversely, lower-order B-spline might inadequately represent sophisticated functional features, potentially resulting in underfitting. Given the specific complexity of the underlying curve for each representative day, we designate the constant B-spline (Order 1) to the Spring curve, the linear B-spline (Order 2) to the Winter curve, and the quadratic B-spline (Order 3) to the Summer and Fall curve.

Figure 4 synthesizes the experimental findings for each representative day, each including four comparison criteria: (a) an evaluation of the knot counts required to meet the same GCV error bound between our introduced ILP algorithms and the equidistant knot method, (b) a comparison of the MaxAE error produced by both algorithms with the identical number of knots, (c) a comparison of the RMSE error produced by these two algorithms using the same number of knots, (d) the resultant fitted curve using the optimal knot distribution.

From these curve approximations, it is consistently observed that our proposed ILP algorithm demands fewer knots to achieve the same GCV error bound as opposed to the equidistant knot method. This superiority becomes particularly pronounced for lower GCV error bound. To illustrate, in Plot (a) for Day_0621 (Summer Day), the minimal GCV error achievable by the equal knot method is 1.92E-09, with 136 knots. This method exhibits challenges in improving upon this accuracy threshold, with any additional knots potentially deteriorating results, likely due to overfitting. In contrast, our ILP algorithm is not constrained by this limitation, managing to achieve a GCV well below 1.92E-09 with fewer knots. For instance, a GCV error of 1.32E-09 was obtained using only 85 knots by the ILP algorithm. Furthermore, from each Plot (b) and (c), it is obvious that the ILP algorithm consistently outperforms in terms of MaxAE and RMSE error with the same number of knots used compared to the equidistant knot methods.

The search for an optimal knot distribution necessitates a careful balance between the model fit and its inherent complexity. To illustrate, for Day_0101(Winter Day), the ILP algorithm attains the minimal MaxAE error of 0.0015 with 83 knots and an RMSE error of 0.00013 using 105 knots. Interestingly, a considerable reduction to merely 27 knots only results in a marginal rise in the MaxAE error to 0.0037 and the RMSE error to 0.00041. This underscores the necessity to trade off approximation precision against the number of basis functions employed. The optimal knot distributions for each representative day are showcased in each Plot (d) of Figure 4. Specifically, Day_0101(Winter Day) employs 58 linear B-spline basis functions, Day_0320(Spring Day) uses 31 constant B-spline basis functions, Day_0621(Summer Day) adopts 51 quadratic B-spline basis functions, and Day_0922 (Fall Day) utilizes 74 quadratic B-spline basis functions.

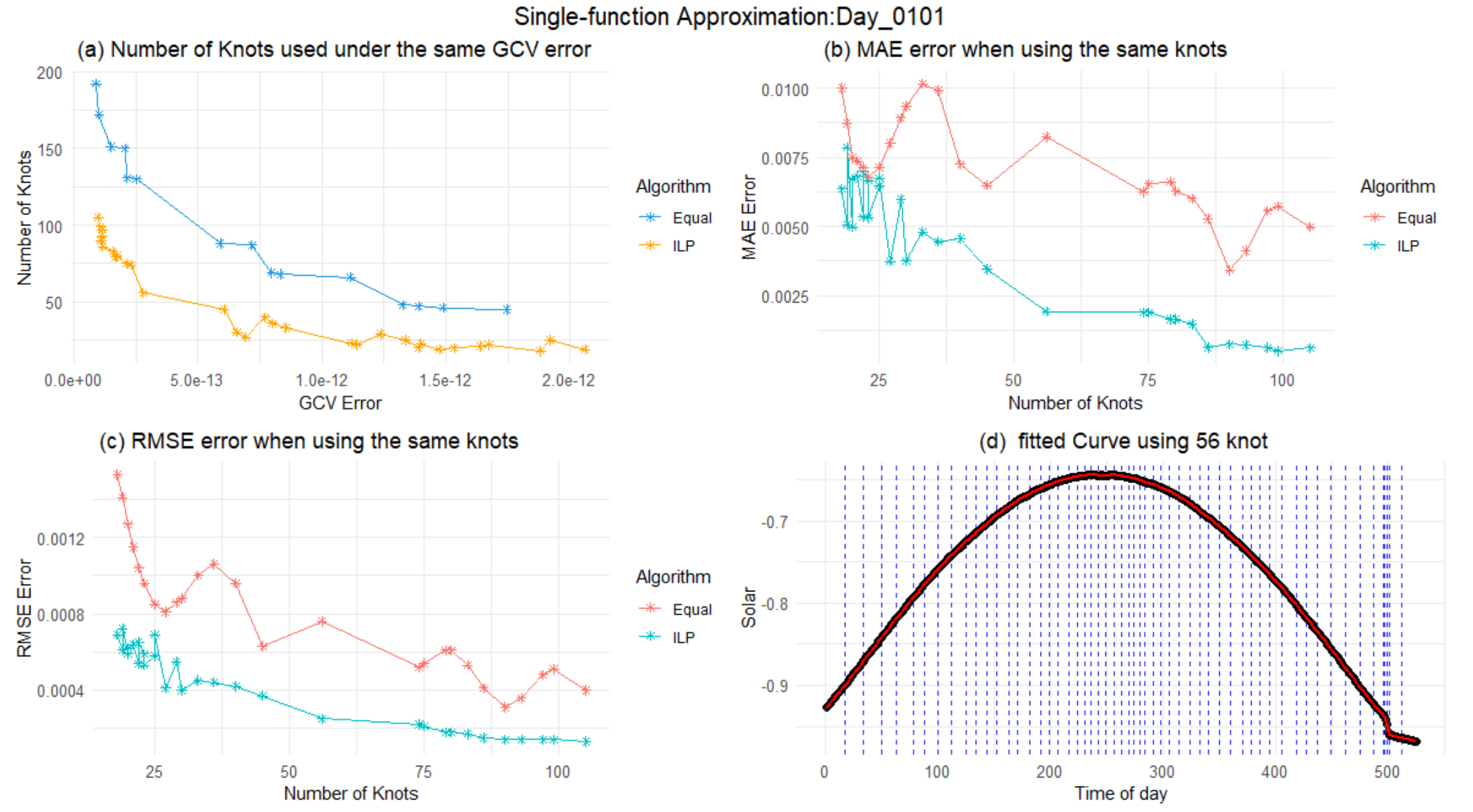
Single-function Approximation:Day_0101
(a) Number of Knots used under the same GCV error
Number of Knots
GCV Error
Algorithm
Equal
ILP
(b) MAE error when using the same knots
MAE Error
Number of Knots
Algorithm
Equal
ILP
(c) RMSE error when using the same knots
RMSE Error
Number of Knots
Algorithm
Equal
ILP
(d) fitted Curve using 56 knot
Solar
Time of day

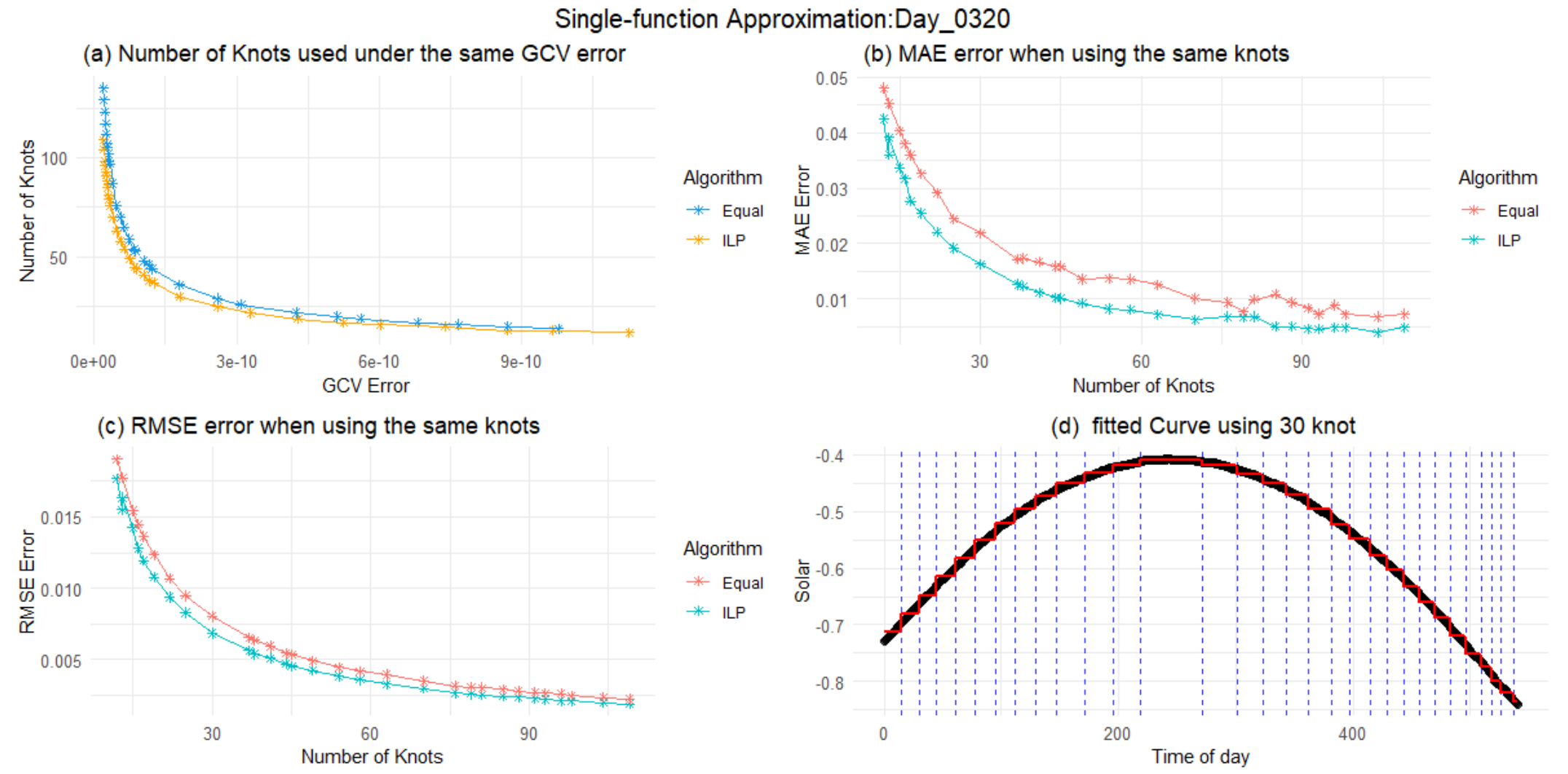
Single-function Approximation:Day_0320
(a) Number of Knots used under the same GCV error
Number of Knots
GCV Error
Algorithm
Equal
ILP
(b) MAE error when using the same knots
MAE Error
Number of Knots
Algorithm
Equal
ILP
(c) RMSE error when using the same knots
RMSE Error
Number of Knots
Algorithm
Equal
ILP
(d) fitted Curve using 30 knot
Solar
Time of day

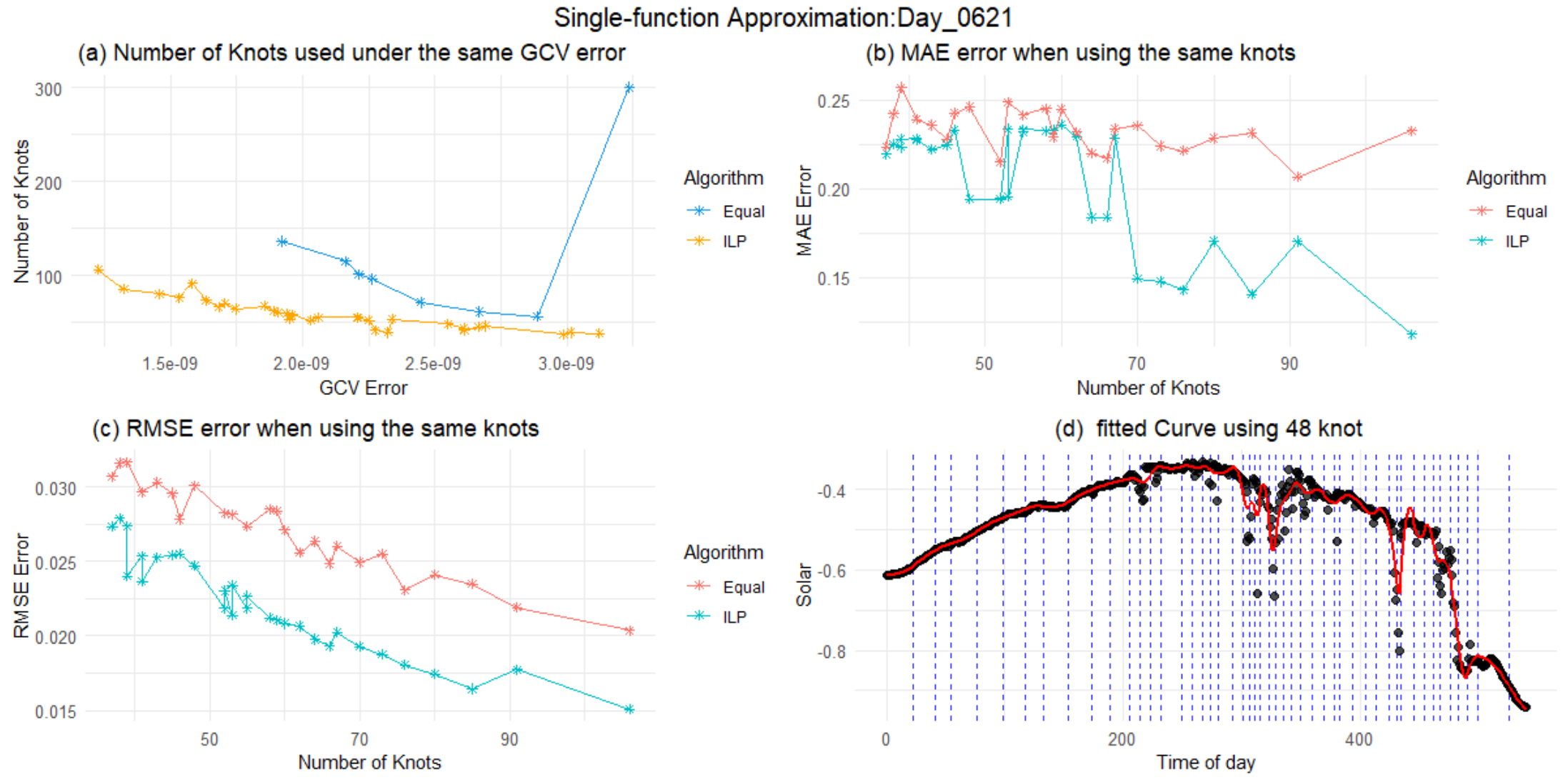
Single-function Approximation:Day_0621
(a) Number of Knots used under the same GCV error
Number of Knots
GCV Error
Algorithm
Equal
ILP
(b) MAE error when using the same knots
MAE Error
Number of Knots
Algorithm
Equal
ILP
(c) RMSE error when using the same knots
RMSE Error
Number of Knots
Algorithm
Equal
ILP
(d) fitted Curve using 48 knot
Solar
Time of day

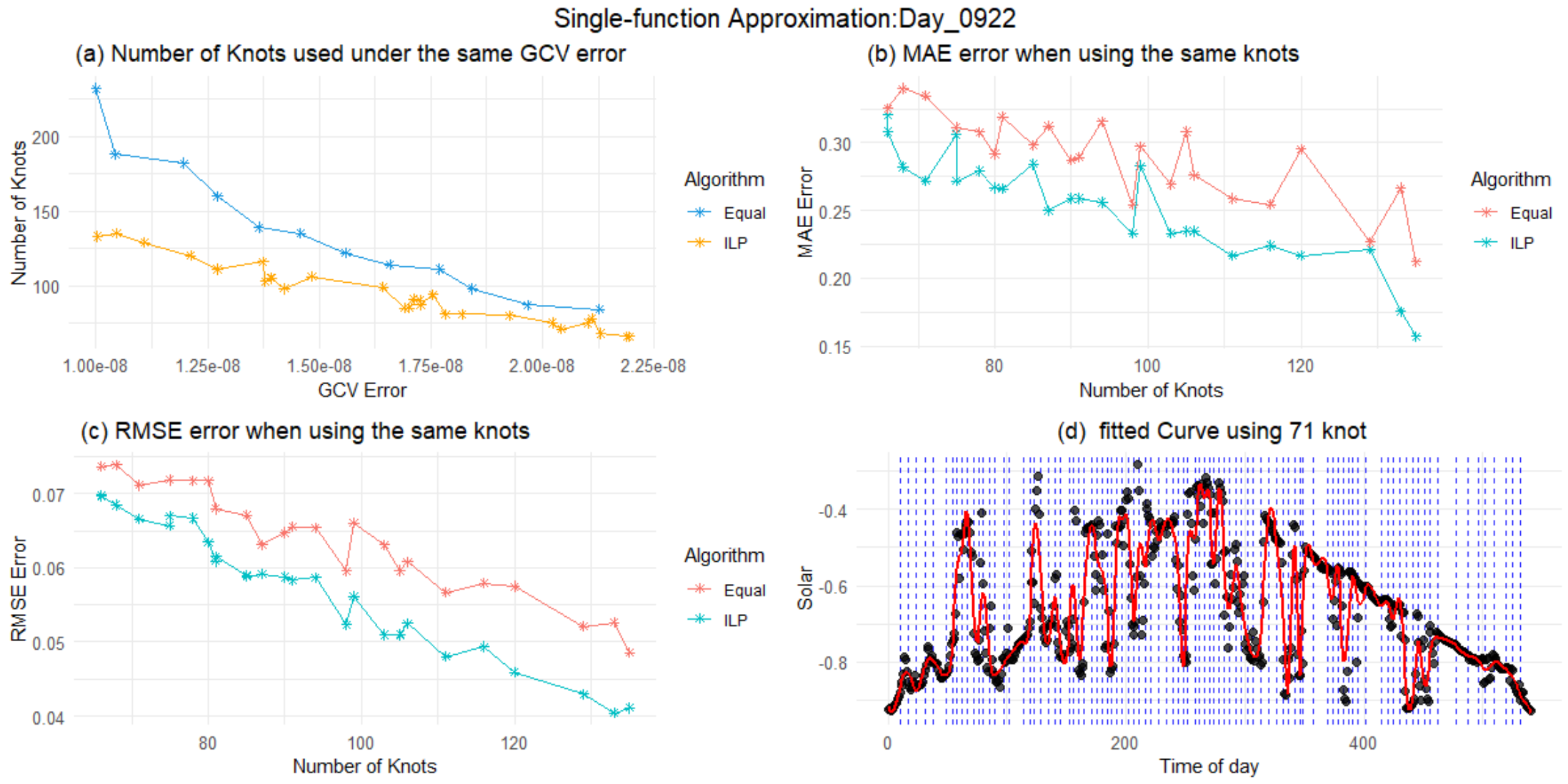

Figure 4. Methods comparison and fitted curve for Four Representative Days: Winter 0101, Spring 0320, Summer 0621, and Fall 0922, respectively.

### 5.1.2. *Unified Multifunction Approximation*

The superior performance of our proposed free knot placement algorithm has been demonstrated in the case of single function approximation, compared to the equidistant knot placement method. However, the task of the unified multifunction approximation introduces greater complications. This is primarily because a common set of B-spline basis functions is used to approximate a set of functions, each of which could potentially exhibit diverse and randomly varying patterns. To assess the efficacy of our proposed method in addressing the unified multifunction approximation problem against the equidistant knot method, we calculate the mean value of MaxAE and RMSE errors across all 366 daily curve approximations, denoted as Mean_MaxAE and Mean_RMSE. Both metrics can serve as the indicators of the overall performance for all day-curve approximations throughout 2016

In our initial analysis, we compare the mean MaxAE and RMSE errors yielded by our proposed ILP algorithm with those of the equidistant knot method when using a consistent number of knots. The comparative results are illustrated in Figure 5. Specifically, Plots (b) and (d) of Figure 5 provide a detailed assessment of mean approximation errors for varying B-spline orders, ranging from Order 1 to Order 4. Our findings reveal that our algorithms consistently outperform the standard equal knot methods in overall approximation performance in terms of both RMSE and MaxAE, regardless of the use of any order of B-spline. An integrated visualization encompassing all B-spline orders is showcased in Plot (a) and (c), with a notable superior performance of the order 3 and 4 B-spline.

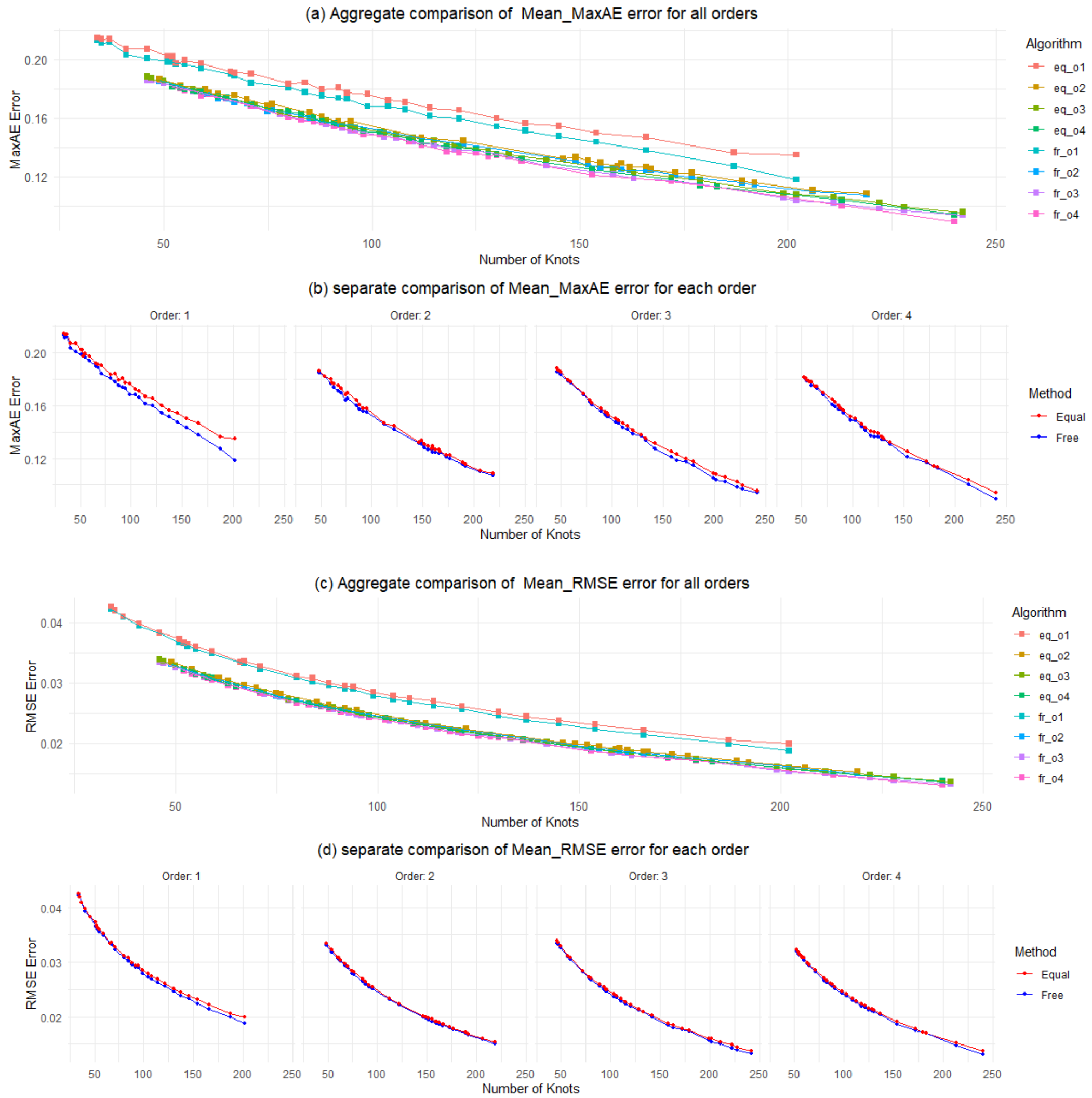

Figure 5: Comparison of mean error using ILP algorithm and Equal knot algorithm for Order 1-4 B-spline.

An alternative metric, the maximal error among all daily function approximations, serves to illustrate the worst approximation scenario. Figure 6 provides a comparison of the Max_RMSE achieved by ILP algorithms and the equal knot method using the same number of knots. The observations from Plot (b) of Figure 6 align with those drawn from the mean error analysis: our algorithms consistently exhibit a lower Max_RMSE in comparison to the conventional equal knot techniques given an identical number of knots for all orders. Moreover, as depicted in Plot (a), B-spline basis functions of orders 2, 3, and 4 manifest analogous performance in terms of Max_RMSE.

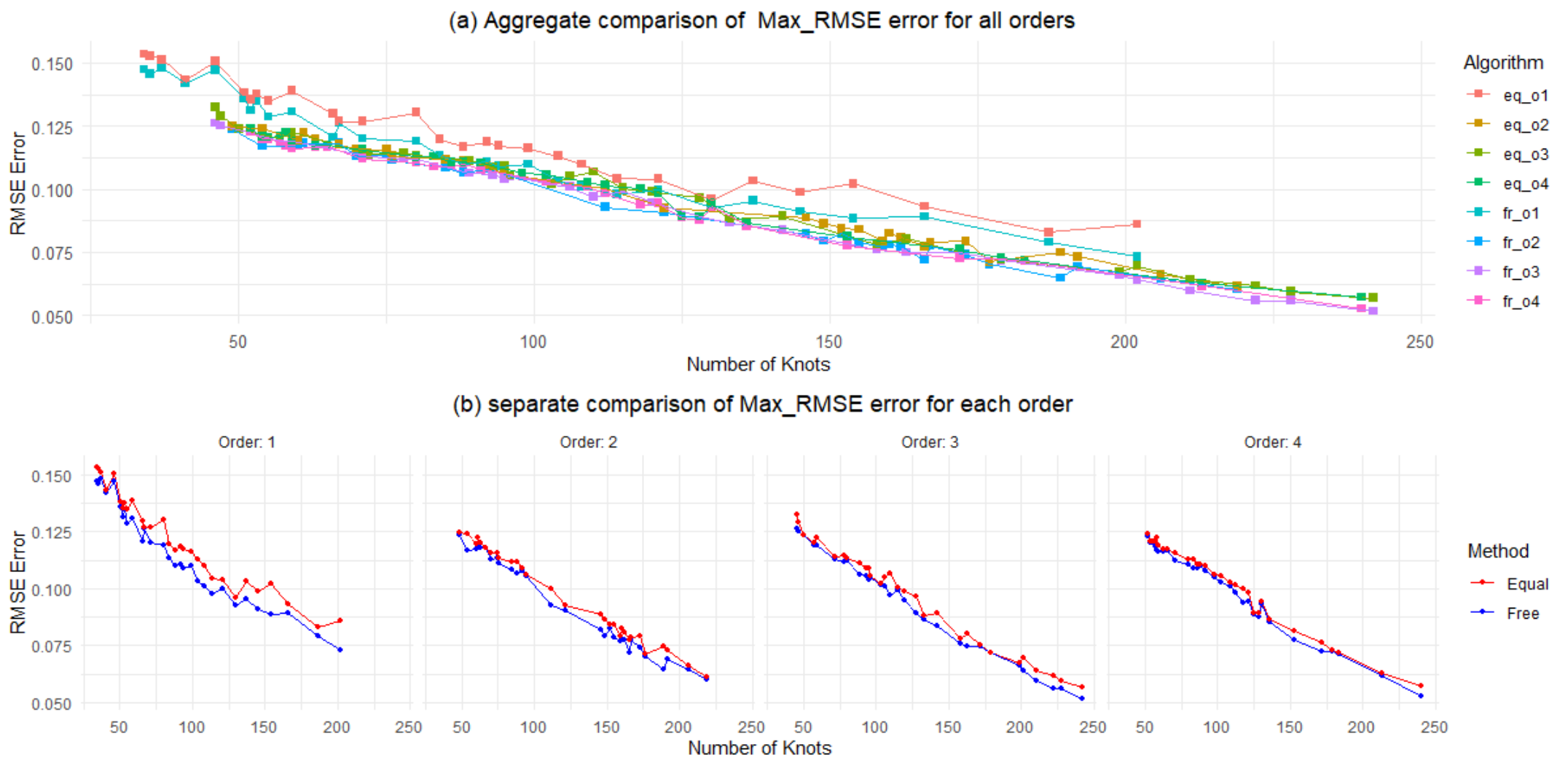

Figure 6: Comparison of Max error using ILP algorithm and Equal knot algorithm for Order 1-4 B-spline.

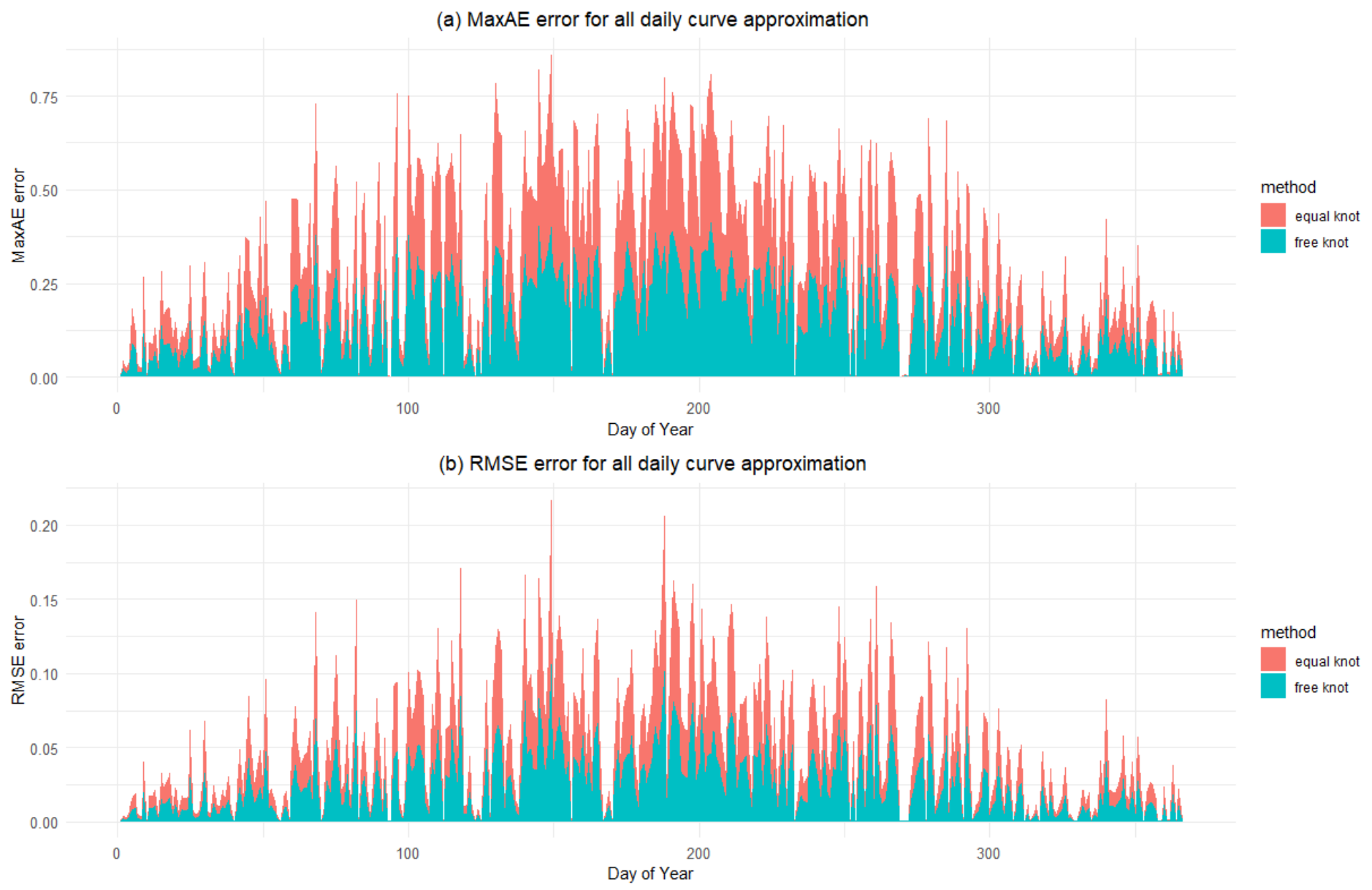

Figure 7: Comparison of RMSE and MaxAE error of each daily curve approximation using the proposed algorithm and Equal method.

Subsequently, the order 3 B-spline basis functions with 89 knots were chosen to approximate all daily functions. As depicted in Figure 7, the ensuing RMSE and MaxAE errors for each daily function approximation based upon such optimal common knot distribution, are markedly lower than their counterparts obtained using the equidistant knot method.

### *5.2. Functional Regression*

In this subsection, we explore the practical application of the proposed M2P functional model across two prediction paradigms: Scalar Output Prediction and Functional Output prediction. Four different experiments were conducted for each paradigm using distinct real-world datasets frequently employed in FDA research. The data for each experiment was partitioned into 75% for training and 25% for testing. We offer a comprehensive comparison between the proposed M2P functional model and three categories of prevailing techniques:

(i) Standard nonfunctional NN models, including the multilayer perceptron (MLP), convolutional neuron network (CNN), long-short-term recurrent (LSTM), and the bidirectional variant of LSTM (Bi-LSTM) (Graves and Schmidhuber 2005).

(ii) Statistical functional regression models, including the functional linear model (FLR), functional principal component (FPC) regression and its variant with roughness penalization (Yao and Müller 2005), functional partial least square (FPLS) regression and its variant with roughness penalization (Preda, Saporta, and Lévéder 2007), and a nonparametric functional model (Ferraty and Vieu 2003).

(iii) NN-based functional model, including the FuncNN model, AdaFNN model mentioned in Section 1, and Functional Basis Neural Network (FBNN) (Rao and Reimher 2023)

For comparison, the conventional MLP architecture was integrated into the M2P model framework, and the tunning of its hyperparameters in the network mirrored the strategy employed for other NN-related models. Except for the AdaFNN model, all functional models necessitate an initial preprocessing to represent discrete measurements in functional data form. An FPC basis function expansion was utilized in comparative functional models for the functional data representation, requiring a priori determination of the optimal number of basis functions. The list of hyperparameters and tuning settings for all methods can be found in Section 3.1 of the supplemental materials. The final model configuration of the neural network and the optimal knot distributions for the functional input or output in the M2P model are provided in Section 3.2 in the supplementary materials. Noteworthy is that some functional methods, such as AdaFNN, are designed only for univariate regression, hence we adopted a single functional variable for consistency. However, the flexibility of our methodology extends its applicability to multivariate contexts as well.

### 5.2.1. Scalar Output Prediction

The main objective of scalar output prediction is to predict the scalar output variable $Y_i$ using the input sequence $(X_{i,1}, X_{i,2}, \ldots, X_{i,J})$. When using functional models, the input sequences were transformed in advance into a functional representation denoted as $X_i(t)$. Alternatively, with nonfunctional models, the sequences were combined into a vector input, denoted as $\boldsymbol{X}_i = [X_{i,1}, X_{i,2}, \ldots, X_{i,J}]$. The performance of these models was assessed via the mean squared prediction error (MSPE) as the benchmark criterion:

$$MSPE = \frac{1}{N_{test}} \sum_{i=1}^{N_{test}} \left(\hat{Y}_i - Y_i\right)^2 \tag{27}$$

where $\hat{Y}_i$ and $Y_i$ represent the predicted and actual value of $i$-th observation of output variable $Y$, and $N_{test}$ indicates the count of subjects within the testing set.

Below is a brief description of the four target datasets and their corresponding experiments:

a) *Tecator data* (Thodbery 1996) contains near-infrared (NIR) spectra and chemical composition measurements of 215 pork samples. The goal of this experiment was to predict the percentage of the fat content of pork using the corresponding NIR spectra measured at 100 wavelengths ranging from 850 to 1050 nm. Moreover, the percentage of water content was used as a scalar predictor for additional information.

b) *Gasoline data* (Kalivas 1997) comprises NIR spectra and octane number of 60 gasoline samples. This experiment aimed to predict the octane number based on the corresponding NIR spectra collected at 400 wavelengths between 900 to 1700 nm.

c) *Bike Sharing data* includes daily bike retail count and 24 hourly temperature values for 102 Saturdays. The objective of the experiment was to predict the square root of the daily bike retail count based on hourly temperatures for a specific day. For more details of the data, see Fanaee-T and Gama 2014.

d) *Diffusion Tensor Imaging* (DTI) data consists of fractional anisotropy (FA) collected at 93 locations along the corpus callous (cca) at 382 visits. The goal of the experiment was to predict the Paced Auditory Serial Addition Test (pasat) score at some visits based on the corresponding FA measurements. We removed 38 visits with missing values. Further details of the data can be seen in Goldsmith et al. (2011).

Table 1. The MSPE errors of the proposed M2P model and all comparison models for four scalar output prediction experiments

| | **Tecator** | **Gasoline** | **Bike Sharing** | **DTI** | |
|---|---|---|---|---|---|
| MLP | 6.70 | 0.29 | 3.54 | 0.86 | Non-Functional Models |
| CNN | 6.27 | 0.37 | 2.89 | 0.89 | |
| LSTM | 4.39 | 0.82 | 2.52 | 0.99 | |
| Bi-LSTM | 4.50 | 0.89 | 2.74 | 1.00 | |
| FLR | 7.20 | 1.28 | 3.37 | 0.85 | Statistical Functional Models |
| FPC regression | 6.50 | 0.78 | 3.23 | 0.87 | |
| FPC regression with 2nd penalization | 6.50 | 0.80 | 4.93 | 0.84 | |
| FPLS regression | 7.38 | 0.91 | 3.37 | 0.95 | |
| FPLS regression with 2nd penalization | 6.93 | 0.96 | 3.07 | 0.87 | |
| Nonparametric regression | 6.87 | 0.88 | 4.95 | 1.04 | |
| FuncNN | 3.12 | 0.09 | 3.45 | 0.77 | NN-based Functional models |
| AdaFNN | N/A | 0.77 | 3.16 | 0.82 | |
| M2P-MLP | **2.80** | **0.04** | **1.71** | **0.72** | Proposed method |

Table 1 presents the final results for all prediction models. As reflected in Table 1, the proposed M2P grounded in the standard MLP architecture outperformed all other models across all four experiments. The NN-based functional model showcased superiority over the statistical function model, possibly due to the power of the neuron network in managing non-linear input-output relationships. Both nonfunctional and statistical functional models demonstrated comparable efficacy throughout all scalar output prediction experiments. Note that in the Tecator experiment, an "NA" result arose due to the limitation of AdaFNN to the univariate scenarios.

### *5.2.2. Functional Output Prediction*

The primary objective of functional output prediction is to predict the functional output $\hat{Y}_i(s)$ based on a given input sequence $(X_{i,1}, X_{i,2}, \dots, X_{i,J})$. Input preprocessing for both functional and non-functional models follows the procedures detailed in the scalar output experiments. Importantly, both SOF functional regression models and non-functional models produce only scalar outputs. To enable a direct comparison across diverse model outputs, the MSPE errors (in Equation 17) were calculated for both SOF functional regression and non-functional models, whose scalar outputs are the mean values of the output sequences. The performance of the FOF functional regression models was evaluated through the Mean Square Prediction Error of the Mean ($MSPE_M$), expressed as

$$MSPE_M = \frac{1}{N_{test}} \sum_{i=1}^{N_{test}} \left( \overline{\hat{Y}_i} - \overline{Y_i} \right)^2 \tag{28}$$

where $\overline{\hat{Y}_i} = \frac{1}{R} \sum_{r=1}^{R} \hat{Y}_i(s_{i,r})$ is the mean of the discretization for the $i$-th predicted function $\hat{Y}_i(s)$ at time steps $\{s_{i,r}\}_{r=1}^{R}$ , and $\overline{Y_i} = \frac{1}{R} \sum_{r=1}^{R} Y_{i,r}$ is the mean of the $i$-th actual output sequence $(Y_{i,1}, Y_{i,2}, \dots, Y_{i,R})$.

Additionally, to comprehensively evaluate the predictive power of the proposed M2P model in the functional output space, another error metric called the Mean Square Prediction Error of the entire function ( $MSPE_F$ ) was employed. This metric quantifies the overall discrepancy between the true function and the predicted function, considering their differences at every point in the domain:

$$MSPE_F = \frac{1}{R * N_{test}} \sum_{i=1}^{N_{test}} \sum_{s=1}^{R} \left(\hat{Y}_{i,s} - Y_{i,s}\right)^2 \tag{29}$$

A concise overview of target datasets and their corresponding experiments is detailed below:

a) Gait data (Ramsay and Silverman 2005) contains hip and knee angles in degrees through a 20-point movement cycle for 39 boys. In this case, we aim to predict a boy's a 20-point gait cycle for his knee angle based on the corresponding cycle for his hip angle.

b) Daily data (Ramsay, Hooker and Graves 2009) comprises 365 daily temperature and precipitation values for 35 Canadian cities. In this case, we aim to predict the annual precipitation curve in a specific city based on its annual temperature curve.

c) Electricity data (UK Power Networks, 2015) includes the hourly electricity consumption of 5015 individual households over two weeks. In this case, we aim to predict a household's weekly electricity consumption curve for the second week based on their first week's electricity curve. Each weekly curve consists of 336 data points.

d) Traffic data, collected from Caltrans performance measurement system (PeMS) consists of 5-minute time resolution traffic occupancy rates on 727 non-holiday workdays. In this case, we aim to predict the occupancy trajectories over T=[10:00, 24:00] using the corresponding trajectories within S=[0:00,10:00]. There are 120 observations in period $T$ and 168 measurements over $S$.

Table 2 presents the $MSPE_M$ values of the FOF functional models with the MSPE errors of the SOF functional and non-functional regression models across the four experiments. The considered FOF models include the linear functional response model (FLR-FOF), the FBNN model, and our proposed M2P model. All other functional models under examination belong to the SOF category. Observations from Table 2 align with previous Scalar Output Prediction experiments, underscoring that the M2P-MLP model outperforms other comparison models across all four datasets. It is anticipated that the FOF models would yield superior outcomes over their SOF counterparts, with models such as FLR-SOF and FLR-FOF as illustrative cases. This is because averaging the output sequence in the SOF models is equivalent to approximating its underlying output function using a consistent B-Spline without any internal knot. Such an approximation inevitably results in substantial information loss and consequently increases prediction errors. Lastly, Figure 8 provides a visual comparison between the $MSPE_F$ values of our proposed M2P model and the FBNN model across the four experiments. The consistently lower the $MSPE_F$ values achieved by the M2P model across all experiments distinctly underline its superior predictive capability not only in estimating the mean value of the function but also the entire function interval.

Table 2. The MSPE-M (or MSPE) errors of the proposed M2P model and all comparison models for four functional output prediction experiments

| | **Gait** | **Daily** | **Electricity** | **Traffic** | |
|---|---|---|---|---|---|
| MLP | 15.85 | 1.64 | 0.0034 | 3.69 | Non-Functional Models |
| CNN | 12.68 | 3.26 | 0.0035 | 4.41 | |
| LSTM | 9.92 | 1.73 | 0.0033 | 3.42 | |
| Bi-LSTM | 9.78 | 1.59 | 0.0031 | 3.21 | |
| FLR-SOF | 10.43 | 2.21 | 0.02016 | 3.66 | Statistical Functional Models |
| FLR- FOF | 9.87 | 1.82 | 0.02020 | 3.65 | |
| FPC regression | 10.33 | 1.08 | 0.0201 | 3.73 | |
| FPC regression with 2nd penalization | 10.24 | 1.21 | 0.0203 | 4.78 | |
| FPLS regression | 9.78 | 1.56 | 0.0216 | 3.56 | |
| FPLS regression with 2nd penalization | 9.70 | 1.57 | 0.0212 | 3.63 | |
| Nonparametric regression | 10.11 | 1.34 | 0.0271 | 4.35 | |
| FuncNN | 9.53 | 1.07 | 0.00311 | 2.90 | NN-based Functional models |
| AdaFNN | 10.98 | 0.95 | 0.00302 | 2.43 | |
| FBNN (FOF) | 7.64 | 1.06 | 0.00305 | 2.61 | |
| M2P-MLP (FOF) | **7.12** | **0.73** | **0.00276** | **2.22** | Proposed method |

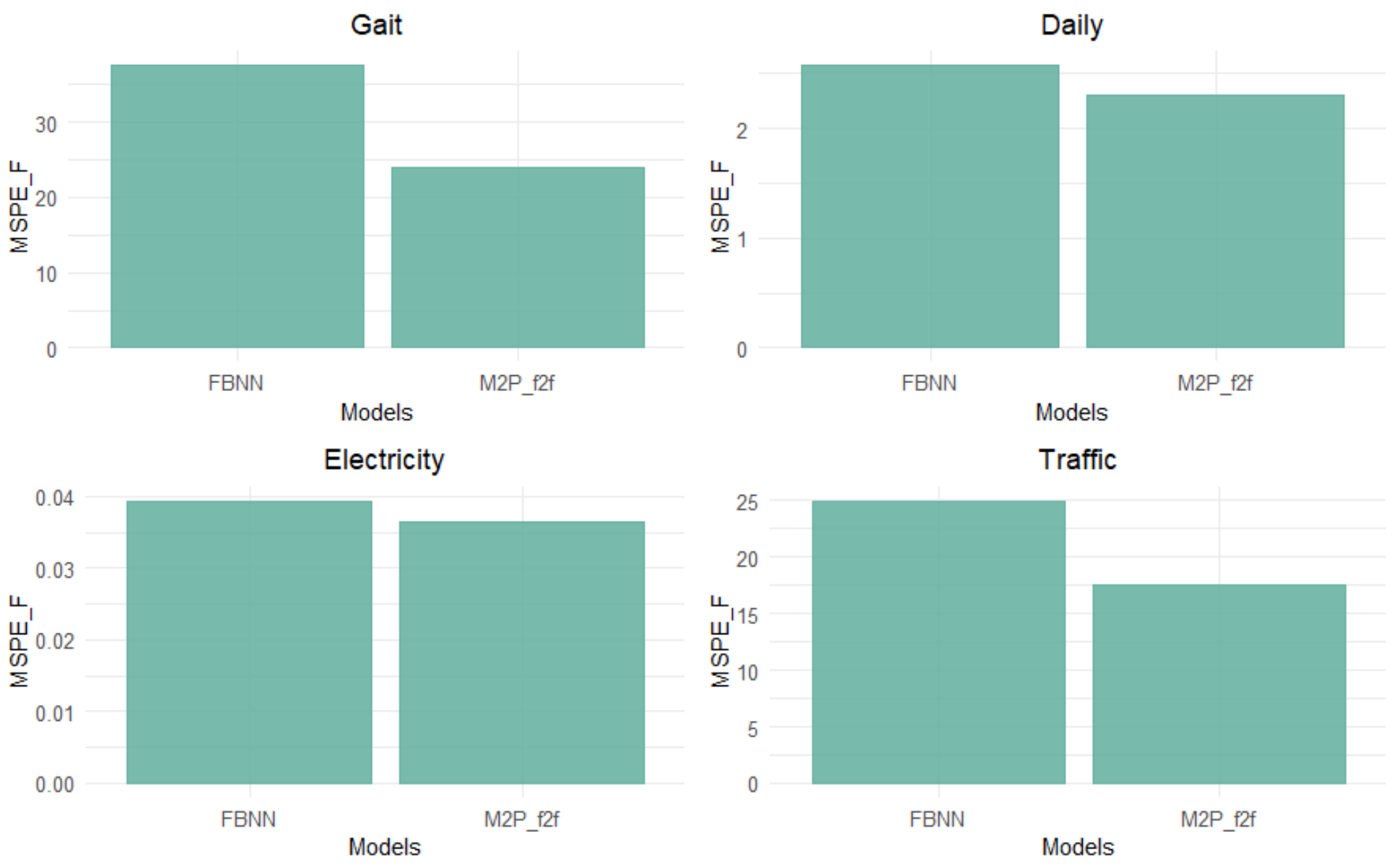

Figure 8. A comparison between the $MSPE_F$ values of the proposed M2P(F2F) model and the FBNN model across the four experiments

## 6. Conclusions and Discussion

This paper introduces a novel approach to nonlinear functional regression by developing the Mapping-to-Parameter (M2P) function model, complemented by the incorporation of free knot B-spline approximation techniques. A new function approximation challenge termed the unified multifunction approximation has been formally defined and tackled. This has been accomplished through the introduction of a novel free knot algorithm devised for the B-spline approximation of multiple functions. The newly proposed iterative local placement approach has demonstrated superior performance over the traditional equidistant knot placement strategy in both tasks of approximating a single daily solar irradiance curve and approximating 366 daily curves.

The proposed M2P model offers a simpler, flexible, and effective alternative to both existing statistical FDA regression models and prevailing neural network-based functional regression models. It introduces a systematical approach to deciding the architecture and configuration of the regression models to replace the trial-and-error approach in the architecture and configuration determination in the NN functional regression model. This obviates the necessity for searching nonlinear relationships in the function space and enables the use of a wide range of supervised learning techniques. In eight real-world experiments, the M2P model consistently demonstrates superior predictive performance in comparison to statistical FDA models, conventional deep learning models, and state-of-the-art FNN models, addressing both scalar output and function output prediction problems.

Despite the promising results, it should be acknowledged that the proposed ILP algorithms determine knot positions by considering the most complex local case among all functions, which may lead to overfitting for some smoother functions. Future research should focus on mitigating this limitation. This could be achieved by developing a local knot modification strategy to eliminate unnecessary knots, or by integrating regularization techniques, such as L1 or L2 penalties. Moreover, there is potential for investigating strategies to handle noisy or sparse data.

## 7. Supplemental Materials

Section 1: The Proof for Lemma 1

Section 2: The Proof for Theorem 1

Section 3: Hyperparameter for All models and tuning setting

## 8. Acknowledgment

The authors report that there are no competing interests to declare. The DTI data were collected at Johns Hopkins University and the Kennedy-Krieger Institute.

# Supplemental Document:

## 1. *Proof of Lemma 1:*

*Let* $p+1 \le m \le n$, *for a function* $f \in W_q^{l+1}(J_m)$ *with* $1 \le q \le \infty, 0 \le l \le p$,

$$J_m := \left[\xi_{m-p-V_L}, \xi_{m+p+1+V_U}\right] \cap [a,b],$$

*where integers* $V_L, V_U \ge -p$,

*According to the theorem 32 of Lyche et.al (2018), we have*

$$\|f - Qf\|_{L_q([\xi_m,\xi_{m+1}])} \le \frac{(2p + V_L + V_U + 1)^{l+1}}{l!}(1 + C)h_{m,\xi}^{l+1}\|D^{l+1}f\|_{L_q(J_m)},$$

*where* $Qf$ *is a B-spline approximation to f in an* $L_q$ *norm, and* $h_{m,\xi}^{l+1}$ *is the largest length of a knot interval in* $J_m$, *and* $C$ *is some constant. However, we do not know the knot sequence in advance in our problem. Then, we choose* $V_L = V_U = -p$ *and for* $J_m = [\xi_m, \xi_{m+1}]$, *the local error bound in the* $L_\infty$ *norm is given as the Equation (15) in the paper:*

$$\|f - Qf\|_{L_\infty([\xi_m,\xi_{m+1}])} \le \frac{h_m^{p+1}}{n!}\|D^{p+1}f\|_{L_\infty([\xi_m,\xi_{m+1}])},$$

*Consider the entire interval* $[a,b]$ *as the union of these subintervals,*

$$[a,b] = \bigcup_{m=p+1}^{n} J_m,$$

*the global approximation error is the maximum of the local error over all subintervals,*

$$\|f - Qf\|_{L_\infty([a,b])} = \max_{m\in[p+1,n]} \|f - Qf\|_{L_\infty(J_m)}.$$

*If there is a constant* $\varepsilon > 0$ *such that each local error is bounded by* $\varepsilon$,

$$\|f - Qf\|_{L_\infty(J_m)} < \varepsilon, \qquad \forall\, m \in [p+1, n].$$

*Then the global approximation error in the max norm is also bounded by* $\varepsilon$,

$$\|f - Qf\|_{L_\infty([a,b])} < \varepsilon.$$

■

## 2. *The Proof of Theorem 1:*

*From Lemma 1 we know that the* $L_\infty$ *gobal error of a degree p B-spline approximation* $Qf_i$ *to a function* $f_i \in C^{n+1}(T)$ *is bounded by* $\varepsilon$ *if only if the condition below is satisfied:*

$$\frac{h_m^{p+1}}{p!} \max_{\xi_m \le t \le \xi_{m+1}} \left| f_i^{(p+1)}(t) \right| < \varepsilon, \forall m.$$

*By (19), the maximum value of the (p+1)-th derivatives of all functions at each* $t \in T$ *is given as*

$$\max_{i \in [1,N]} \left| f_i^{(p+1)}(t) \right| \le C_t.$$

*Then, we have*

$$\frac{h_m^{p+1}}{p!} \max_{\xi_m \le t \le \xi_{m+1}} \left| f_i^{(p+1)}(t) \right| \le \varepsilon \frac{h_m^{p+1}}{p!} \max_{\xi_m \le t \le \xi_{m+1}} |C_t|, \forall m.$$

*With the stated assumptions for all* $m \in [p+1, n]$, *we have*

$$\frac{h_m^{p+1}}{p!} \max_{\xi_m \le t \le \xi_{m+1}} |C_t| < \varepsilon,$$

*the global bound for all function by* $\varepsilon$ *in the max norm follows immediately.*

■

## 3. *Hyperparameter for all models and tunning setting*

The common hyperparameter in the network includes the Number of Layers, Number of Neurons per Layer, Epochs, validation split, learning rate, the activation function per layer, batch size, and early stopping patience. There are some other hyperparameters for specific NN-related models. For CNN, we focused on CNN 1 with a kernel size of two and filters of 32. For AdaFNN, each neuron of the basis layer is a micro neural network, thus the number of layers and the number of neurons in the micro neural network are the additional hyperparameters for AdaFNN. In addition, two levels of regularization are also included in the AdaFNN model and the orthogonal regularization penalty and the L1 regularization penalty need also to be tuned. For functional models, the choice of basis function and the number of basis functions for functional data are typical hyperparameters. For FuncNN, another hyperparameters are the choice of functional weight basis function and Functional weight basis expansion size. For our proposed M2P model, the choice of the free knot algorithm and the error threshold are two hyperparameters in the functional approximation process, and the common hyperparameter in the network is applied in the prediction process.

The tunning approach is to take a list of possible values for each parameter and run a 5-fold cross-validation. We consider the number of layers between 2 to 5, the number of Neurons in [16,512], the Epochs in [100, 500], the validation split rate in [0.01, 0.2], the learning rate in the range [0.001, 0.1], the choice activation function among {“relu”, “tanh”,” linear”}, batch size in [16, 256] and early stopping patience in [10,20]. For AdaFNN, the orthogonal regularization penalty is among {0, 0.5, 1}, whereas the L1 penalty is among {0，1，2}. For the functional model, the choice of basis function is the FPC basis function, and the number of basis functions is determined in advance, as shown in Table A.1

Table A. The choice of basis function and the number of FPC basis functions for the functional comparison model

| Dataset | Functional Input | Functional Output |
|---|---|---|
| Tecator | 29 | |
| Gasoline | 65 | |
| Bike Sharing | 31 | |
| DTI | 27 | |
| Gait | 13 | 11 |
| Daily | 65 | 20 |
| Electricity | 65 | 65 |
| Traffic | 21 | 31 |

For the M2P model, the final optimal knot distribution and the maximum derivative curve is shown in Figure below and final model configuration for each dataset are listed on the bottom of the figure.

Figure A. The optimal knot distribution (order 2 B-spline with 12 knots) and the maximum derivative curve for **Tecator Data.**

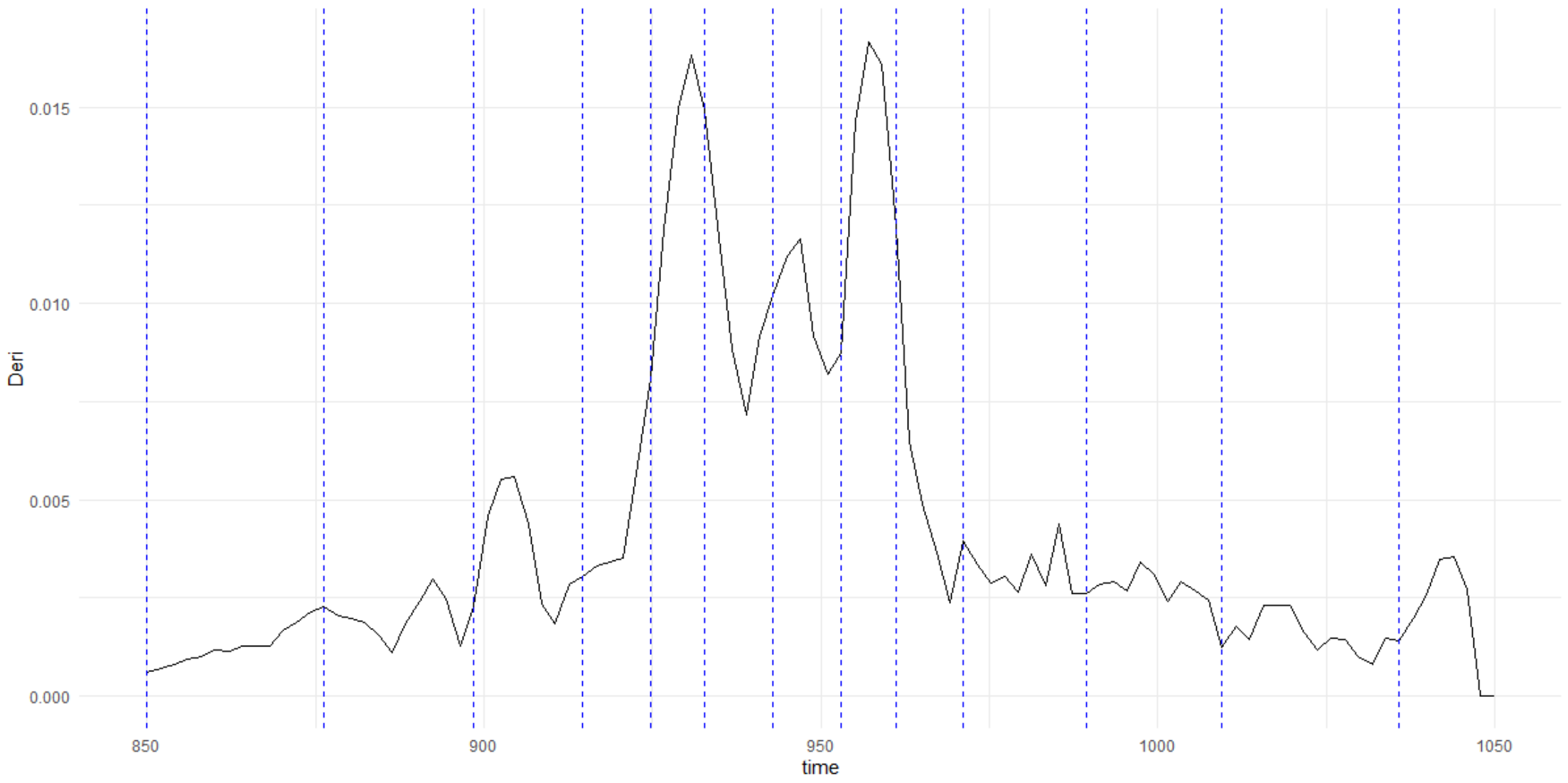

Table B. The final model configuration for **Tecator Data**

| No. Neuron /layer | Activation function /layer | Learn rate | Epoch | Validation split | Batch size | Early stopping |
|---|---|---|---|---|---|---|
| 512,256,128,1 | Relu, Relu, Relu, Linear | 0.005 | 300 | 0.2 | 32 | 10 |

Figure B. The optimal knot distribution (order 3 B-spline with 18 knots) and the maximum derivative curve for **Gasoline Data**.

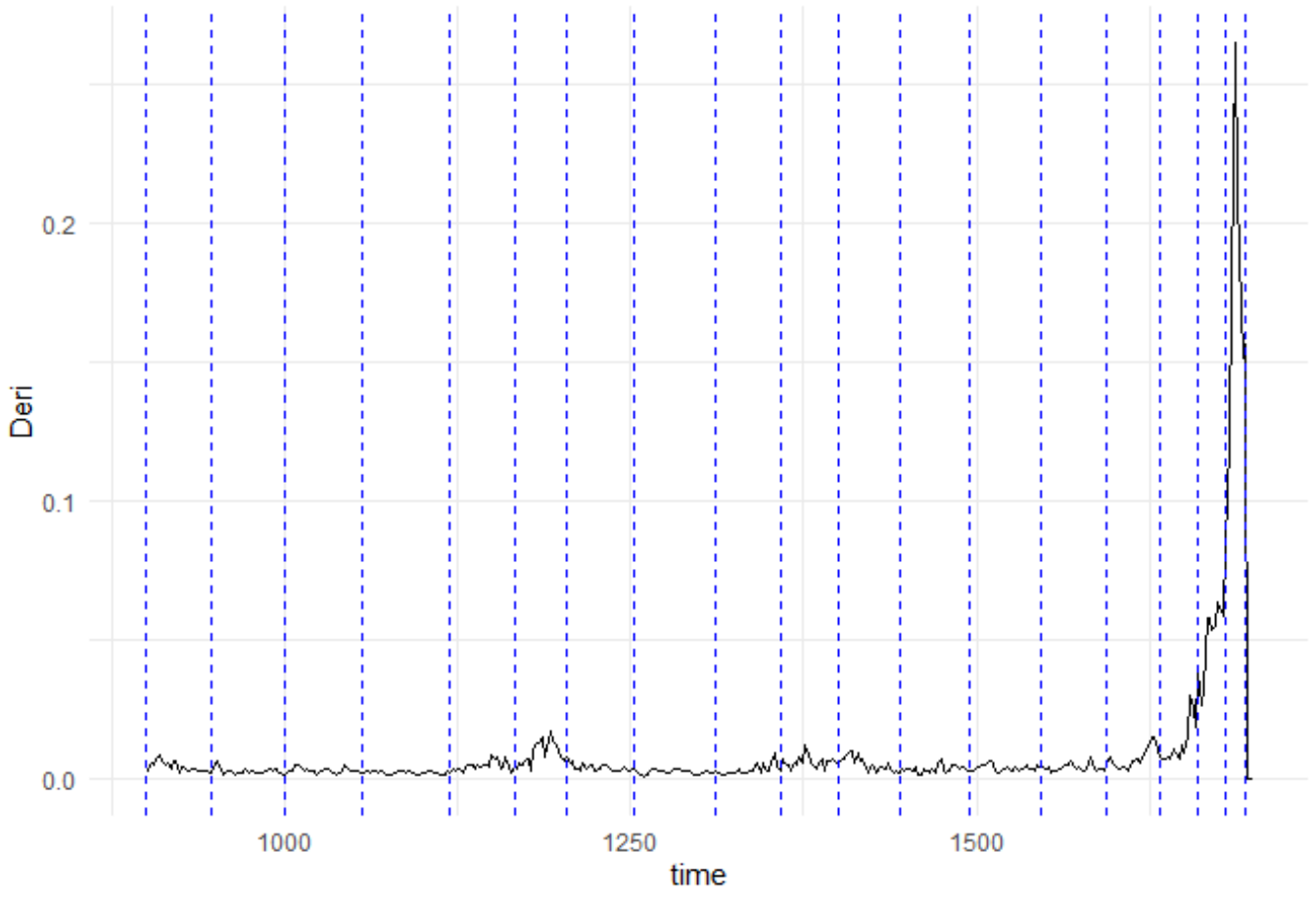

Table C. The final model configuration for **Gasoline Data**

| No. Neuron /layer | Activation function /layer | Learn rate | Epoch | Validation split | Batch size | Early stopping |
|---|---|---|---|---|---|---|
| 64,64,64,1 | Relu, Relu, Relu, Linear | 0.002 | 500 | 0.2 | 32 | 20 |

Figure C. The optimal knot distribution (order 2 B-spline with 8 knots) and the maximum derivative curve for **Bike Data.**

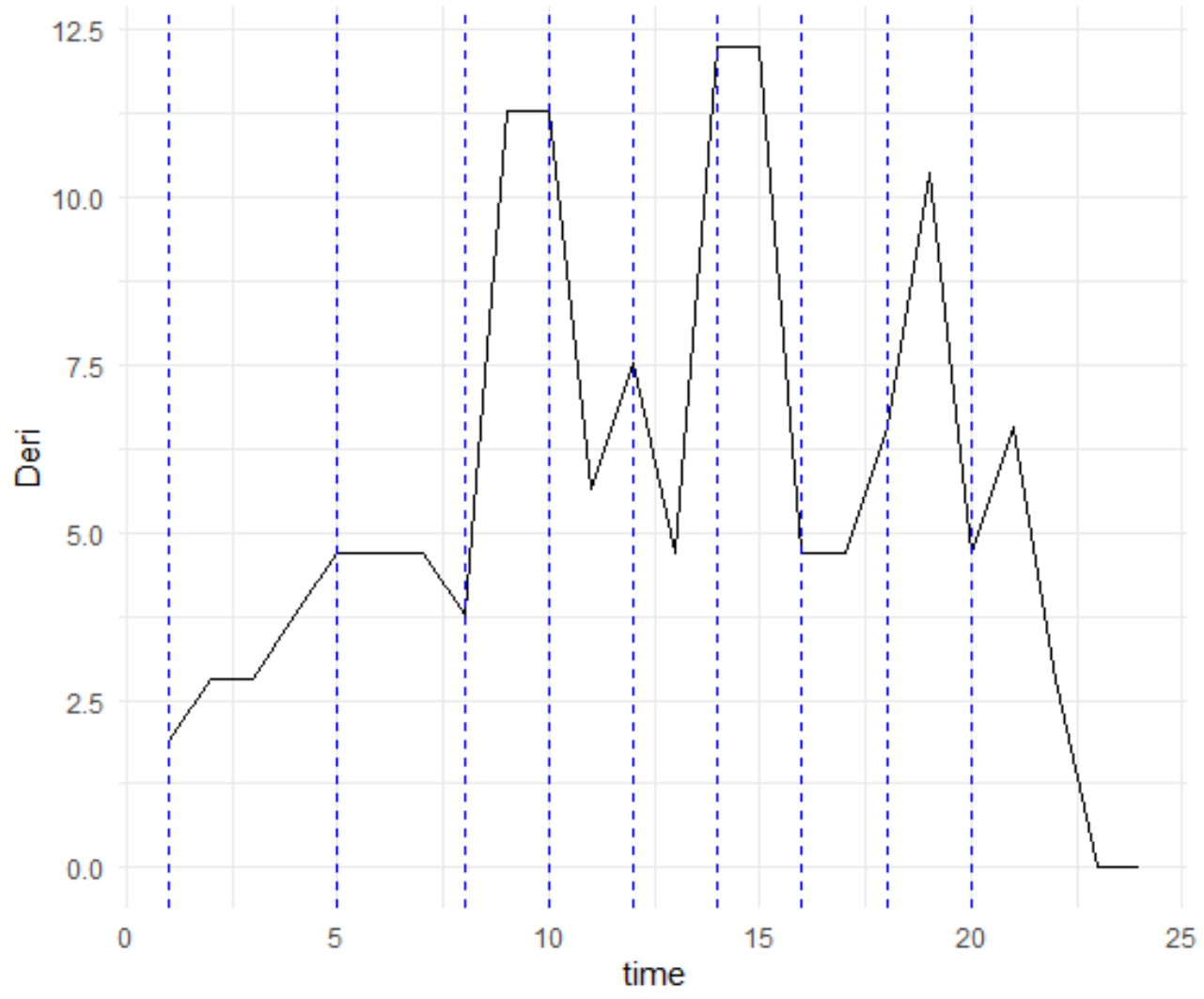

Table D. The final model configuration for **Bike Data**

| No. Neuron /layer | Activation function /layer | Learn rate | Epoch | Validation split | Batch size | Early stopping |
|---|---|---|---|---|---|---|
| 128,32,32 ,1 | sigmoid, sigmoid, Relu, Linear | 0.002 | 500 | 0.15 | 32 | 15 |

Figure D. The optimal knot distribution (order 1 B-spline with 22 knots) and the maximum derivative curve for **DTI Data**.

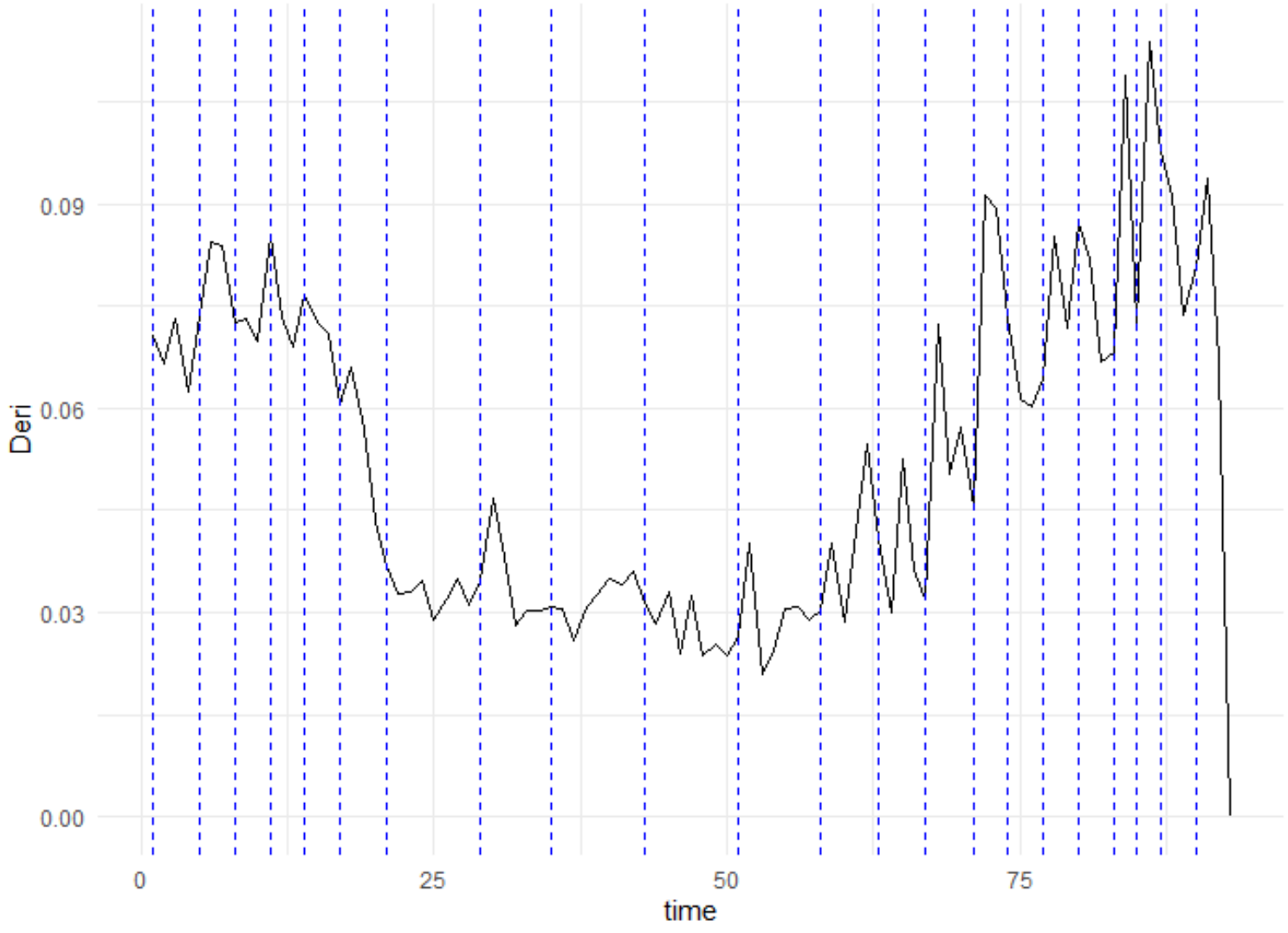

Table E. The final model configuration for **DTI Data**

| No. Neuron /layer | Activation function /layer | Learn rate | Epoch | Validation split | Batch size | Early stopping |
|---|---|---|---|---|---|---|
| 256,256,256,256,256,1 | Relu, Relu, Relu, Relu, Relu , Linear | 0.001 | 500 | 0.2 | 32 | 20 |

Figure E. The optimal knot distribution and the maximum derivative curve for **Gait Data**. (1). Knee: order 4 B-spline with 6 knots (2). Hip: order 3 B-spline with 7 knots determined by ILP_pw algorithm.

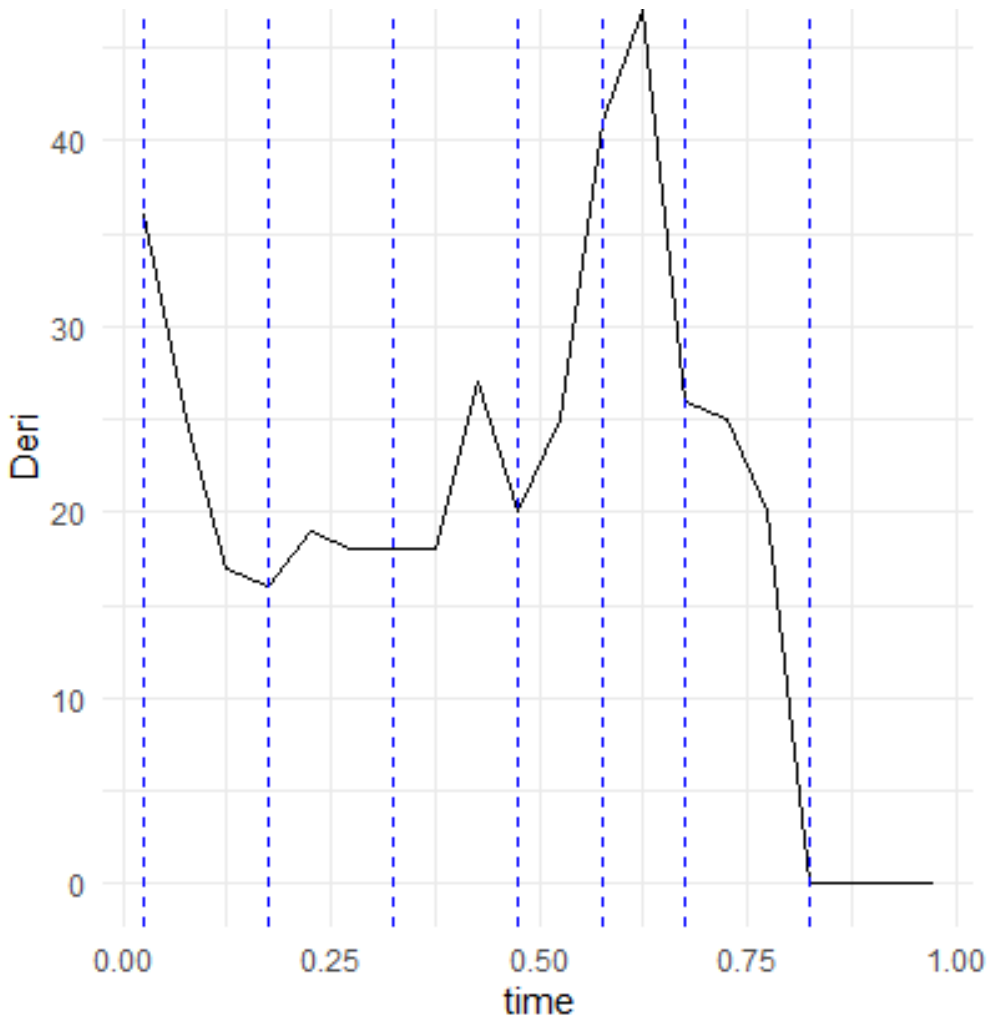

(1)

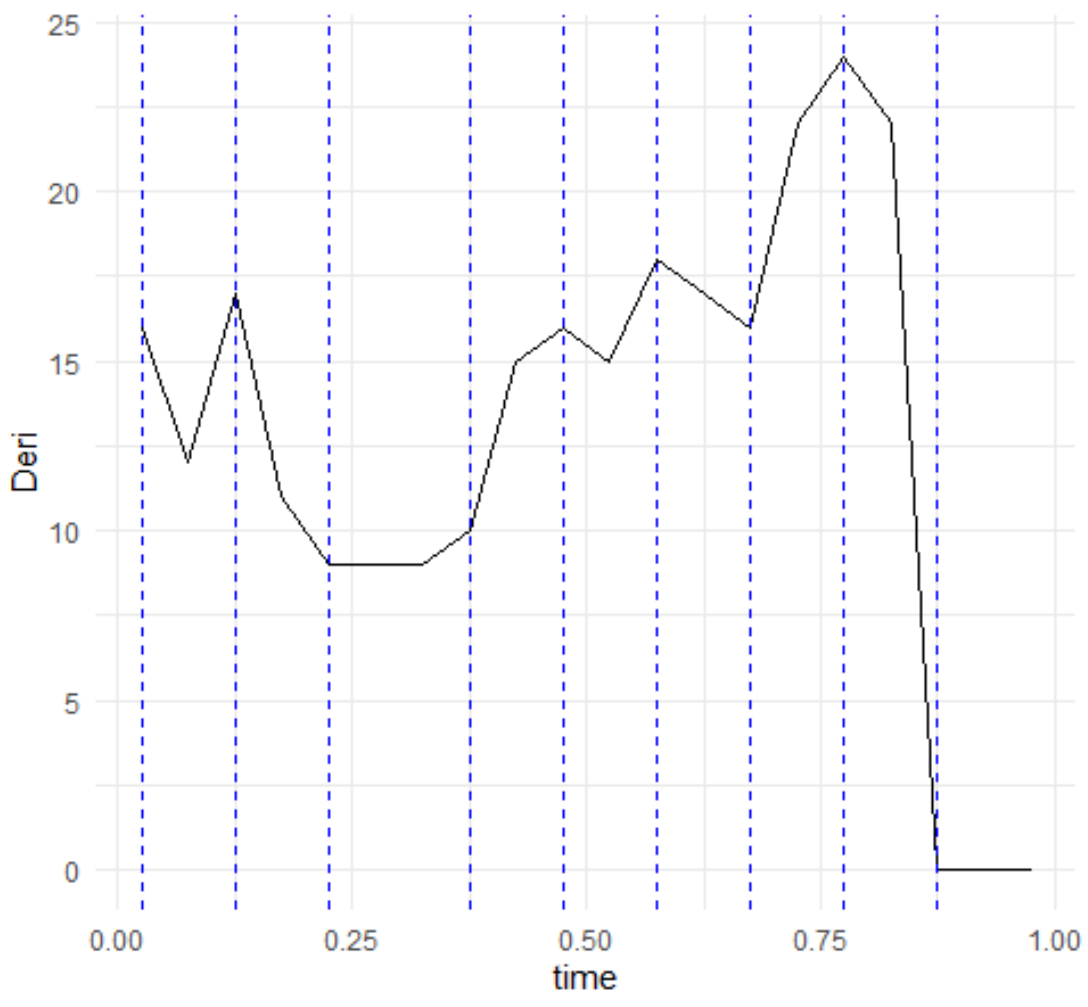

(2)

Table F. The final model configuration for **Gait Data**

| No. Neuron /layer | Activation function /layer | Learn rate | Epoch | Validation split | Batch size | Early stopping |
|---|---|---|---|---|---|---|
| 32,32,32,32,32,1 | sigmoid, sigmoid , Relu, Relu, Relu , Linear | 0. 1 | 300 | 0.2 | 32 | 10 |

Figure F. The optimal knot distribution and the maximum derivative curve for **Daily Data**. (1). Daily Prec: order 2 B-spline with 60 knots (2). Daily Temp: order 1 B-spline with 33 knots

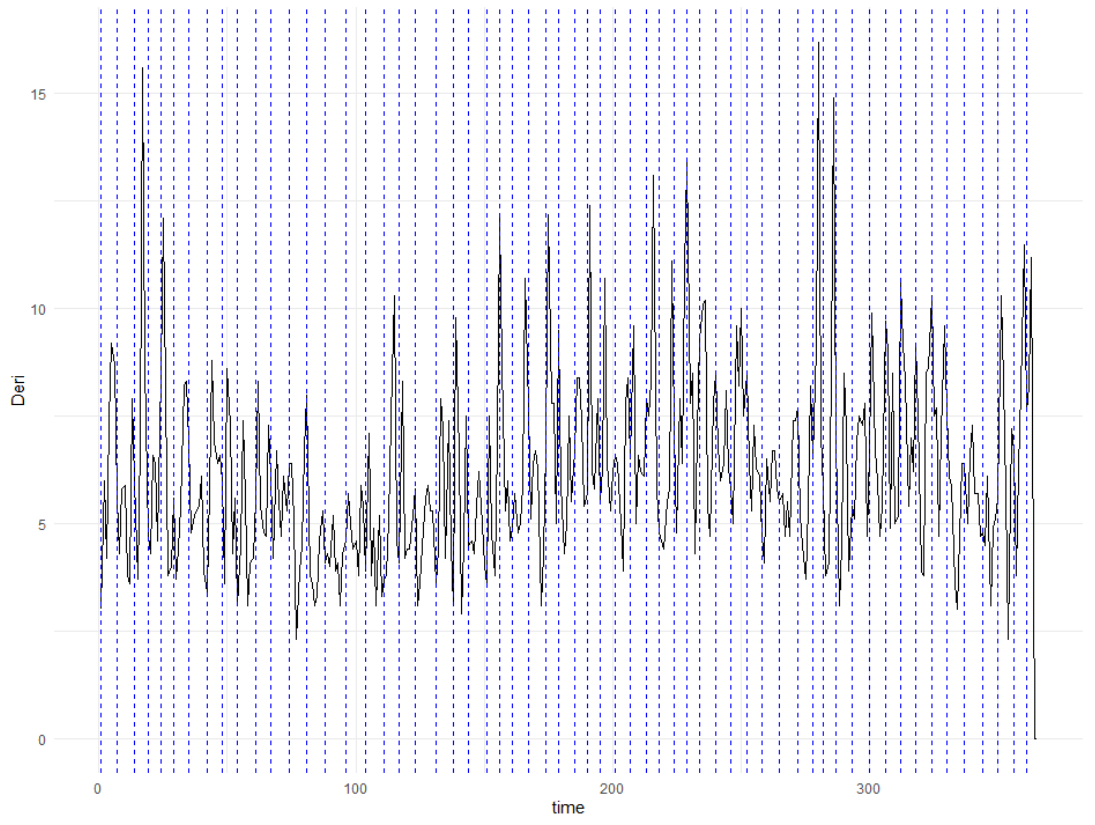

(1)

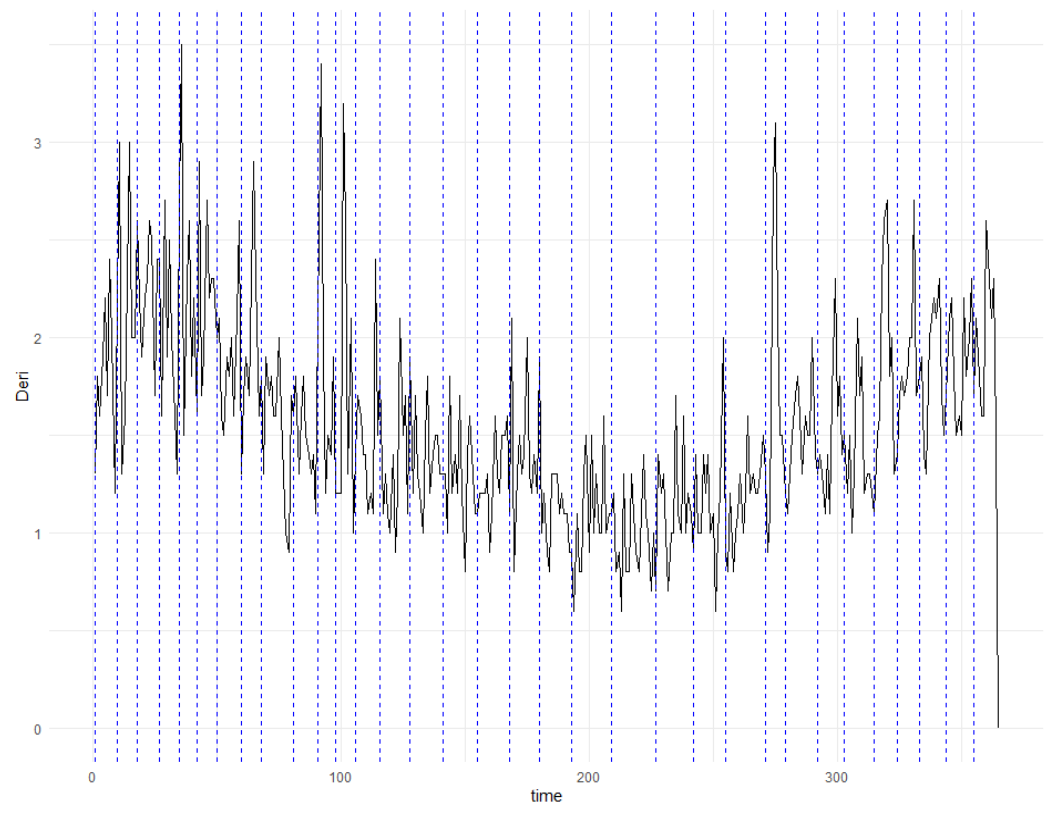

(2)

Table G. The final model configuration for **Daily Data**

| No. Neuron /layer | Activation function /layer | Learn rate | Epoch | Validation split | Batch size | Early stopping |
|---|---|---|---|---|---|---|
| 256,128,128 | Tanh, tanh, tanh, Linear | 0. 001 | 300 | 0.2 | 32 | 20 |

Figure G. The optimal knot distribution and the maximum derivative curve for **Electricity Data**. (1). Week 1: order 3 B-spline with 25 knots (2). Week2: order 2 B-spline with 22 knots

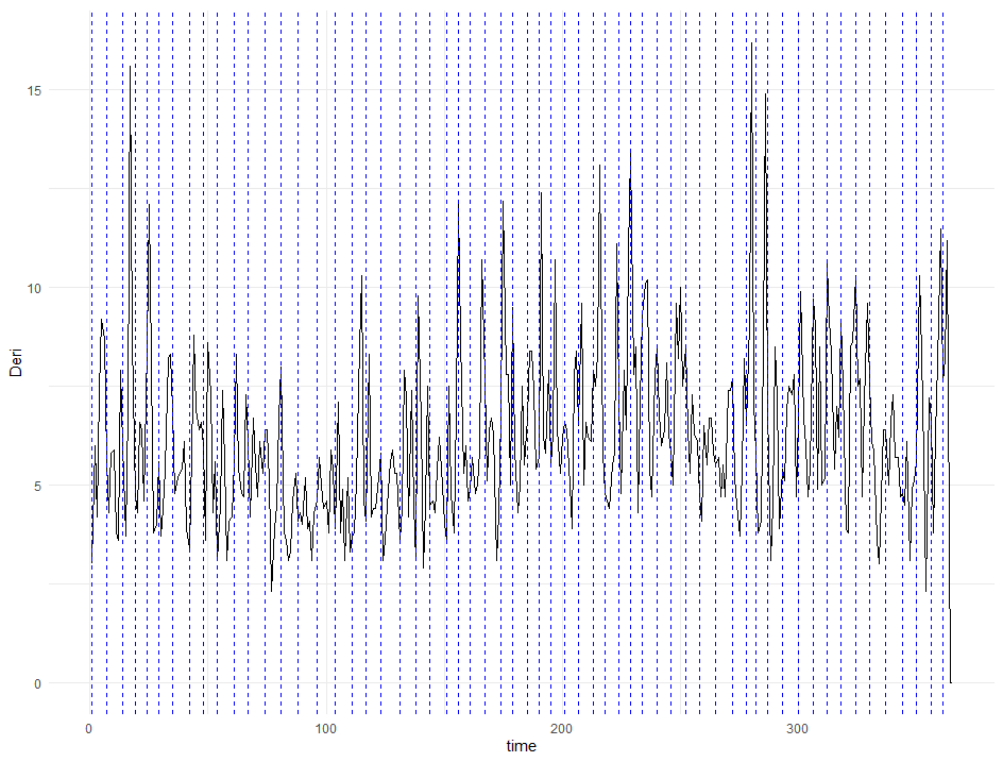

(1)

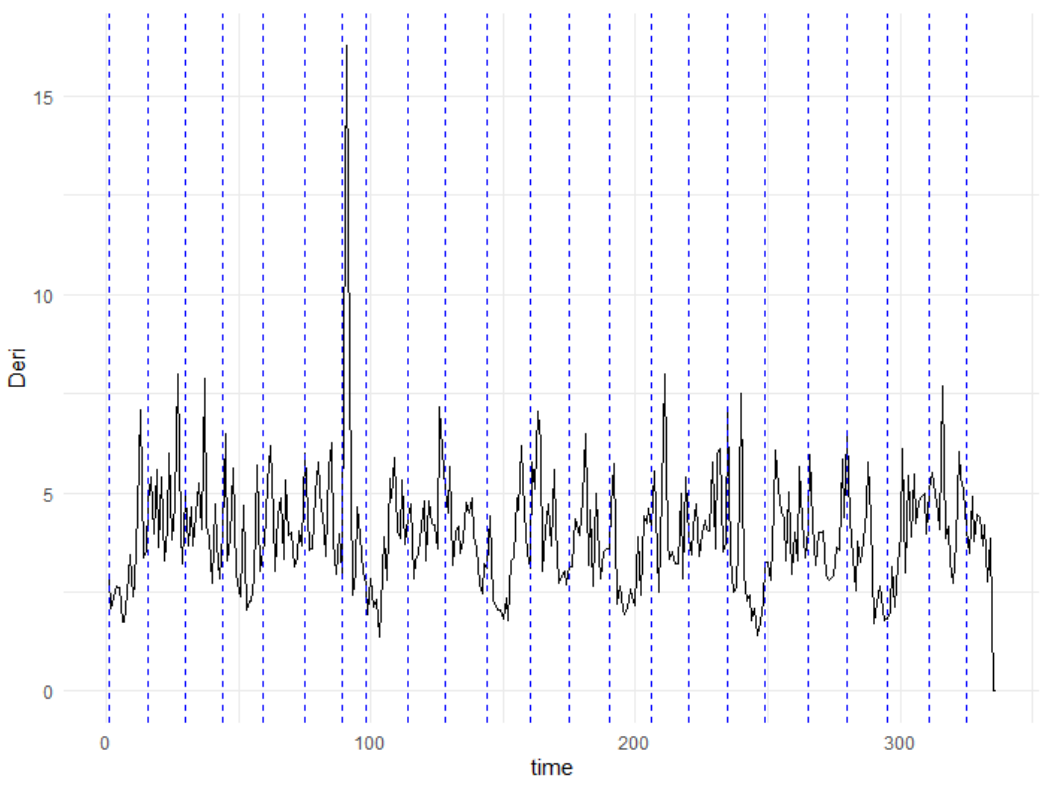

(2)

Table H. The final model configuration for **Electricity Data**

| No. Neuron /layer | Activation function /layer | Learn rate | Epoch | Validation split | Batch size | Early stopping |
|---|---|---|---|---|---|---|
| 32,32,32,32,32 | Relu, Relu, Relu, Relu, Relu, Linear | 0. 001 | 500 | 0.2 | 32 | 20 |

Figure H. The optimal knot distribution and the maximum derivative curve for **Traffic Data**. (1). X: order 2 B-spline with 44 knots; (2). Y: order 2 B-spline with 82 knots

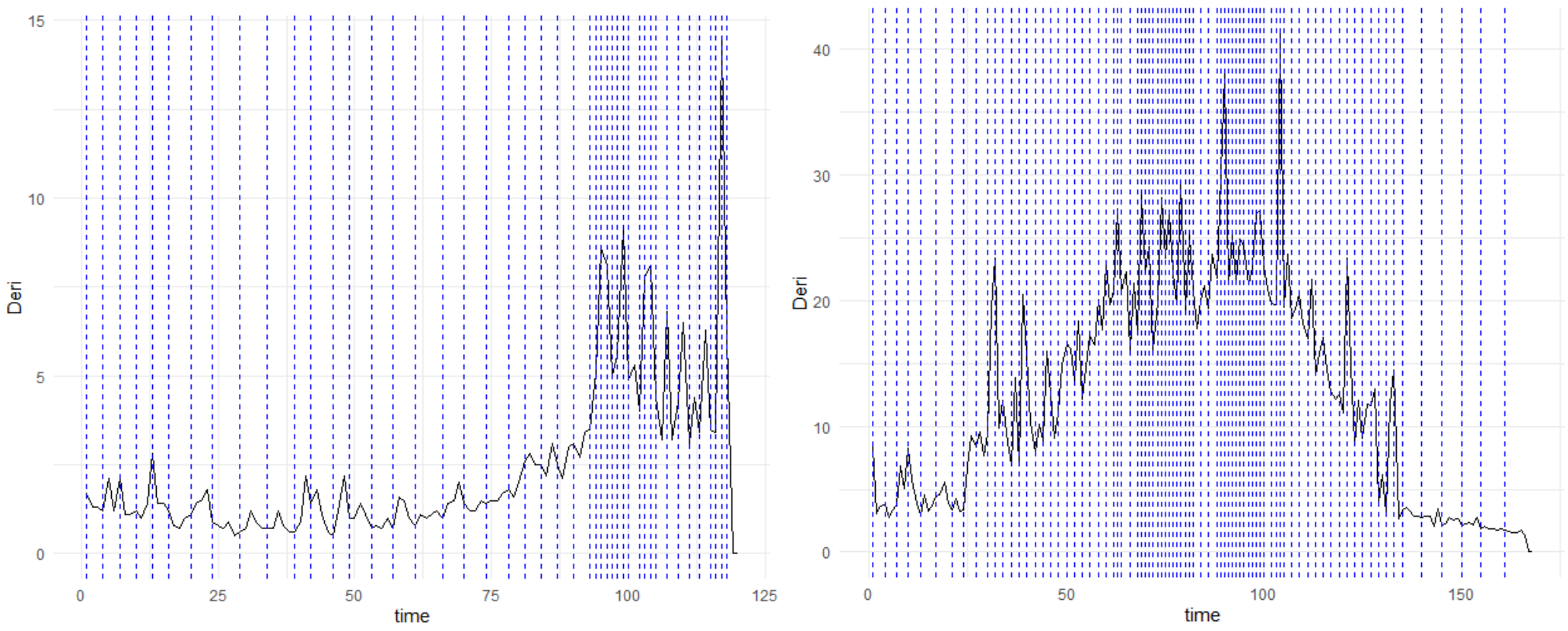

(1) (2)

Table I . The final model configuration for **Traffic Data**

| No. Neuron /layer | Activation function /layer | Learn rate | Epoch | Validation split | Batch size | Early stopping |
|---|---|---|---|---|---|---|
| 16,16,61 | relu, relu, relu, | 0. 001 | 500 | 0.2 | 32 | 20 |